\PassOptionsToPackage{table,dvipsnames}{xcolor}
\documentclass[10pt,twocolumn,letterpaper]{article}

\usepackage{iccv}      

\usepackage{colortbl}  
\usepackage[pagebackref,breaklinks,colorlinks,allcolors=iccvblue]{hyperref}

\usepackage{amsmath,amssymb,amsthm,amsfonts}
\usepackage{bm}
\usepackage{fix-cm}
\usepackage{enumitem}
\newtheorem{assumption}{Assumption}
\usepackage{multirow}

\newtheorem{theorem}{Theorem}

\newtheorem{remark}{Remark}
\newtheorem{proposition}{Proposition}
\newtheorem{corollary}[theorem]{Corollary}

\usepackage[accsupp]{axessibility} 
\usepackage{url}
\usepackage{booktabs}
\usepackage{makecell}
\usepackage{tabularx}
\usepackage{graphicx}
\usepackage[linesnumbered, ruled, vlined]{algorithm2e}

\usepackage{float}
\allowdisplaybreaks[4]  
\newcommand{\gain}[1]{\textcolor{red}{(+#1)}}

\definecolor{lightgray}{gray}{0.9}
\definecolor{red}{rgb}{1,0,0}
\definecolor{Gray}{gray}{0.88}
\definecolor{iccvblue}{rgb}{0.21,0.49,0.74}
\def\paperID{6153} 
\def\confName{ICCV}
\def\confYear{2025}

\title{TokenUnify: Scaling Up Autoregressive Pretraining for Neuron Segmentation}

\author{
Yinda Chen$^{1,2}$\thanks{Equal Contribution.}\hspace{0.5em}
Haoyuan Shi$^{1,2}$\footnotemark[1]\hspace{0.5em}
Xiaoyu Liu$^{1}$\hspace{0.5em}
Te Shi$^{2}$\\
Ruobing Zhang$^{3,2}$\hspace{0.5em}
Dong Liu$^{1}$\hspace{0.5em}
Zhiwei Xiong$^{1,2}$\thanks{Project Leader}\hspace{0.5em}
Feng Wu$^{1,2}$\thanks{Corresponding Author}\\
\\
$^{1}$MoE Key Laboratory of Brain-inspired Intelligent Perception and Cognition,\\
\quad University of Science and Technology of China\\
$^{2}$Institute of Artificial Intelligence, Hefei Comprehensive National Science Center\\
$^{3}$Institute for Brain and Intelligence, Fudan University\\
\texttt{\small \{cyd0806, haoyuan.shi\}@mail.ustc.edu.cn, \{zwxiong, fengwu\}@ustc.edu.cn}
}

\begin{document}
\maketitle

\begin{abstract}
Neuron segmentation from electron microscopy (EM) volumes is crucial for understanding brain circuits, yet the complex neuronal structures in high-resolution EM images present significant challenges. EM data exhibits unique characteristics including high noise levels, anisotropic voxel dimensions, and ultra-long spatial dependencies that make traditional vision models inadequate. Inspired by autoregressive pretraining in language models, we propose TokenUnify, a hierarchical predictive coding framework that captures multi-scale dependencies through three complementary learning objectives. TokenUnify integrates random token prediction, next-token prediction, and next-all token prediction to create a comprehensive representational space with emergent properties. From an information-theoretic perspective, these three tasks are complementary and provide optimal coverage of visual data structure, with our approach reducing autoregressive error accumulation from $O(K)$ to $O(\sqrt{K})$ for sequences of length $K$. We also introduce a large-scale EM dataset with 1.2 billion annotated voxels, offering ideal long-sequence visual data with spatial continuity. Leveraging the Mamba architecture's linear-time sequence modeling capabilities, TokenUnify achieves a 44\% performance improvement on downstream neuron segmentation and outperforms MAE by 25\%. Our approach demonstrates superior scaling properties as model size increases, effectively bridging the gap between pretraining strategies for language and vision models. Code is available at \url{https://github.com/ydchen0806/TokenUnify}.
\end{abstract}    
\begin{figure*}[t]
    \centering
\includegraphics[width=0.9\linewidth]{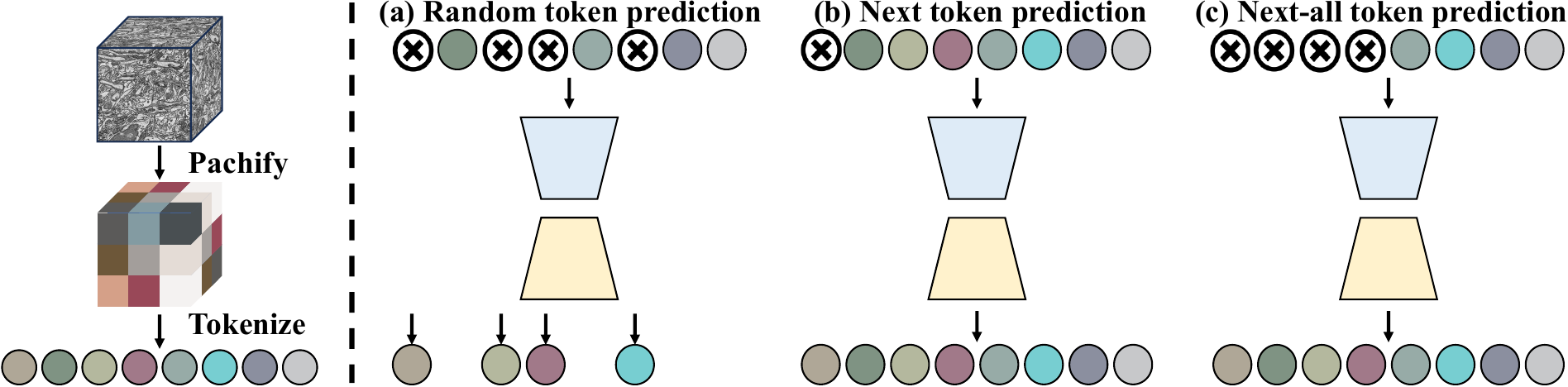}
    \caption{TokenUnify divides the 3D EM image into non-overlapping patches, which are tokenized into a sequence of K tokens. Three complementary tasks are performed for rich 3D image representations: (a) random token prediction for spatial pattern learning, (b) next token prediction for sequential dependencies, and (c) next-all token prediction for global structure modeling.}
    \label{fig:teasor}
\end{figure*}

\section{Introduction}
Understanding neural circuits is a central goal in neuroscience. Electron microscopy (EM) imaging enables reconstruction of neurons and synaptic connections at high resolution~\cite{lynn2019physics, ohno2015three}. However, EM neuron segmentation presents significant challenges due to three key properties: (1) high noise levels from electron beam interactions, (2) anisotropic voxel dimensions with coarser z-axis resolution, and (3) long-range spatial continuity spanning thousands of patches. Current methods struggle to capture global context and long-range dependencies~\cite{chen2023self, huang2022learning, liu2024cross, sheridan2023local}. Despite success in natural language processing~\cite{liu2023pre}, similar advancements for long-sequence visual data remain largely unexplored.

Large language models (LLMs) show remarkable scaling capabilities~\cite{achiam2023gpt, touvron2023llama, touvron2023llama2} through effective autoregressive modeling. However, visual data has more complex structures than text. Contrastive methods like DINO v2~\cite{oquab2023dinov2} and masked reconstruction like MAE~\cite{chen2023self, he2022masked} demonstrate strong representation learning but lack favorable scaling laws~\cite{singh2023effectiveness}. Masked prediction faces theoretical limitations where estimation error scales as $O(\sqrt{s \log p/n})$ for sparsity $s$, dimensionality $p$, and sample size $n$. Autoregressive approaches like AIM~\cite{el2024scalable} and LVM~\cite{bai2023sequential} attempt to bridge this gap, but error accumulation ($O(K)$ with sequence length $K$) and computational complexity hinder their application to high-resolution EM images~\cite{bachmann2024pitfalls}.

This paper tackles visual autoregressive learning challenges through hierarchical predictive coding. Let $X = \{x_1, x_2, \ldots, x_K\}$ represent visual tokens from an EM image, where $\mathcal{M} \subset \{1, \ldots, K\}$ denotes masked indices and $\mathcal{M}^c$ denotes unmasked tokens. We recognize that three prediction tasks capture complementary visual structure aspects: random token prediction learns noise-robust spatial patterns, next-token prediction captures sequential dependencies, and next-all token prediction models global context. Under regularity conditions (AR($\infty$) process with summable coefficients), autoregressive models achieve asymptotically optimal prediction error $O(\sigma^2)$ as capacity increases, where $\sigma^2$ is irreducible noise.

We propose TokenUnify, a unified framework that integrates these three predictive tasks within a coherent information-theoretic formulation (Fig.~\ref{fig:teasor}). This creates a comprehensive representational space where random token prediction provides local robustness, next-token prediction ensures structural continuity, and next-all token prediction captures global morphology, enabling simultaneous encoding of fine-grained details and global relationships unachievable with single objectives. Our approach reduces cumulative errors from $O(K)$ to $O(\sqrt{K})$ by distributing prediction errors across multiple positions rather than sequential accumulation, analogous to central limit theorem scaling. To handle massive high-resolution EM data efficiently, we leverage the Mamba architecture, which reduces computational complexity from quadratic to linear while maintaining superior performance compared to Transformers. We validate our approach on a large-scale EM dataset collected from mouse brain slices, featuring thousands of continuous tokens that ensure spatial continuity crucial for understanding complex neuronal structures. Through comprehensive annotation of six functional brain regions totaling 1.2 billion voxels, we create one of the largest manually annotated EM neuron segmentation datasets to date.

Comprehensive evaluation demonstrates that TokenUnify achieves a 44\% improvement on downstream neuron segmentation and outperforms MAE~\cite{he2022masked} by 25\% in pretraining performance. Furthermore, TokenUnify demonstrates superior scaling properties, offering a promising paradigm for pretraining large-scale visual models.

Our contributions can be summarized as follows:
\begin{enumerate}
    \item We introduce a novel hierarchical predictive coding framework that unifies three distinct visual structure perspectives—random token prediction for spatial robustness, next-token prediction for sequential continuity, and next-all token prediction for global context—within a coherent information-theoretic formulation. Our theoretical analysis (detailed in supplementary material) demonstrates why these prediction tasks are complementary and together provide optimal coverage of visual data structure, while reducing autoregressive error accumulation from $O(K)$ to $O(\sqrt{K})$.
    
    \item We achieve a 44\% performance improvement on downstream neuron segmentation tasks and realize the first billion-parameter Mamba network for visual autoregression, demonstrating both effectiveness and computational efficiency in processing long-sequence visual data. Our approach outperforms existing methods including MAE by 25\% while exhibiting favorable scaling properties as model size increases.
    
    \item We construct one of the largest EM neuron segmentation datasets with 1.2 billion finely annotated voxels across six functional brain regions, providing an ideal testbed for long-sequence visual modeling. This comprehensive dataset offers spatially continuous visual data with thousands of tokens per volume, enabling rigorous validation of autoregressive methods for biological image analysis.
\end{enumerate}
\section{Related Work}
\label{sec:relatedwork}

\paragraph{Neuron Segmentation.}
In the field of EM image analysis, neuron instance segmentation has emerged as a vital task. Turaga et al. \cite{turaga2010convolutional} pioneered the use of convolutional neural networks (CNNs) for affinity generation, followed by clustering of voxels in the affinity graph using post-processing techniques such as waterz \cite{funke2018large} and LMC \cite{beier2017multicut}. Since then, affinity-based methods have seen significant advancements. Funke et al. \cite{funke2018large} introduced the MALIS loss \cite{briggman2009maximin} during training to ensure topologically accurate segmentation. Recent work by Huang et al. \cite{huang2022learning} proposed an embedding pyramid module to capture affinity information across different scales. Liu et al. \cite{liu2022biological} combined both affinity and embedding features with graph neural networks (GNNs) to enhance the segmentation of adjacent objects in the feature space. Furthermore, Liu \cite{liu2024cross} introduced a knowledge distillation framework, transferring knowledge from 3D models to 2D models, which resulted in improved performance and efficiency in neuron segmentation. Despite these advances, mamba-based architectures \cite{gu2023mamba}, have yet to be fully explored for the anisotropic resolution characteristics of EM imaging. In this paper, we enhance the Segmamba decoder \cite{xing2024segmamba} to better process the anisotropic features of EM data using an affinity-based approach.

\vspace{-0.5cm}
\paragraph{Self-supervised Pretraining.}
Self-supervised pre-training has been a cornerstone for enhancing model representation capabilities. Approaches based on contrastive learning for representation extraction \cite{chen2024learning,zbontar2021barlow,he2020momentum,li2021consistent} and masked reconstruction methods \cite{he2022masked,chen2023self,li2023mage,ding2022unsupervised,chen2024bridging,chen2023point} have shown promise. However, current vision models have not exhibited the same sublinear scaling laws as language models. To address this issue, some tasks have adopted autoregressive pre-training paradigms similar to those used in language models \cite{chen2020generative,bai2023sequential,el2024scalable}, though the training costs remain a significant concern. In this paper, we explore the potential of long visual token autoregressive pre-training and introduce a collaborative training scheme for long token prediction. Our approach aims to balance the scaling laws and training costs, demonstrating improvements in fine-grained visual tasks.
\section{Theoretical Motivations}
\label{sec:theory}
This section provides the theoretical foundations that motivate our approach. Detailed derivations and proofs are provided in the supplementary material.

\textbf{Notation.} Let $\mathbf{X} = \{x_1, x_2, \ldots, x_K\}$ represent a sequence of visual tokens extracted from an EM image, where $\mathcal{M} \subset \{1, \ldots, K\}$ denotes masked token indices, $\mathcal{M}^c$ denotes unmasked tokens, and $x_{<i} = \{x_1, \ldots, x_{i-1}\}$ represents tokens preceding position $i$.

Our framework is motivated by two key theoretical insights: (1) limitations of masked prediction approaches in high-dimensional spaces, and (2) the complementary nature of different prediction tasks from an information-theoretic perspective.

For high-dimensional visual data like EM images, we can demonstrate that conventional masked autoencoding methods face scaling limitations. Specifically, under standard sparsity assumptions, the estimation error of masked prediction scales as $O(\sqrt{s \log p/n})$, where $p$ is the data dimensionality, $s$ is the sparsity, and $n$ is the sample size. This indicates diminishing returns when scaling model capacity for complex visual data.

In contrast, autoregressive models can achieve asymptotically optimal prediction error as model capacity increases under appropriate regularity conditions. For an AR($\infty$) process with summable coefficients, the one-step prediction error converges to the irreducible noise level $\sigma^2$ as model order increases. Our analysis shows that combining three distinct prediction strategies maximizes the total information extracted from visual data:
\begin{equation}
\mathcal{I}_{\text{total}} = \mathbb{E}[I(x_i; x_{\mathcal{M}^c})] + I(x_i; x_{<i}) + I(\{x_i, \ldots, x_K\}; x_{<i}),
\label{eq:total_info}
\end{equation}
where $I(\cdot; \cdot)$ denotes mutual information, and each term represents the information captured by random token prediction, next-token prediction, and next-all token prediction, respectively. Importantly, our next-all prediction strategy reduces error accumulation from $O(K)$ to $O(\sqrt{K})$ for sequences of length $K$, providing natural regularization against error propagation.

\textbf{Latent Manifold Structure.}
The integration of multiple prediction objectives induces a structured latent representation space. For the learned manifold $\mathcal{M}_\theta$, we characterize its sectional curvature $\kappa_\theta$ as:
\begin{equation}
\kappa_\theta(v_1, v_2) \approx 
\begin{cases}
0 \pm \epsilon & \text{if } v_1, v_2 \in \mathcal{T}_{\text{local}} \\
-c & \text{if } v_1, v_2 \in \mathcal{T}_{\text{global}}
\end{cases},
\label{eq:curvature}
\end{equation}
where $\mathcal{T}_{\text{local}}$ and $\mathcal{T}_{\text{global}}$ denote the tangent subspaces encoding local and global features respectively, $\epsilon > 0$ is a small constant, and $c > 0$ is the magnitude of negative curvature. This geometric property enables efficient encoding of both fine-grained details (in near-flat regions) and complex hierarchical structures (in hyperbolic-like regions).


\section{Method}
\subsection{Overview}
\begin{figure*}[t]
    \centering
    \includegraphics[width = \linewidth]{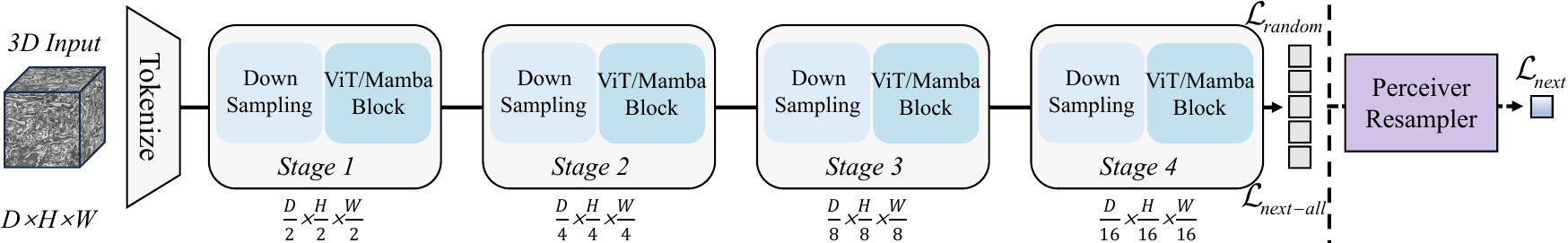}
    \caption{Illustration of the TokenUnify pretraining workflow. The image $\mathbf{X}$ is fed into the Tokenizer, transforming it into a long sequence of tokens $x_i|_{i=1}^K$. The three prediction tasks (random token, next token, and next-all token) are performed with complementary objectives. The Perceiver Resampler is employed to convert varying-size large feature maps into a few visual tokens.}
    \label{fig:framework}
\end{figure*}

The workflow of our method is shown in Fig. \ref{fig:framework}. Our approach comprises two stages: (1) hierarchical predictive pre-training, where we learn a generic visual representation $f_{\theta_1}(\cdot)$ from unlabeled data $\mathbf{X}$ by integrating three complementary prediction tasks that capture both local and global dependencies; and (2) task-specific fine-tuning, where we adapt the pre-trained model to downstream tasks using both $\mathbf{X}$ and labels $Y$ through a task-specific model $g_{\theta_2}(\cdot)$ initialized with $\theta_1$. For EM neuron segmentation, we process input volumes $\mathbf{X} = \{X^{(1)}, \ldots, X^{(T)}\}$, where each $X^{(t)} \in \mathbb{R}^{D \times H \times W}$ represents a 3D image, by partitioning them into smaller patches $x \in \mathbb{R}^{D' \times H' \times W'}$ and leveraging Mamba's efficient sequence modeling capabilities to handle the high-dimensional, long-sequence nature of the data.

\subsection{Hierarchical Predictive Coding in Token Space}
\label{sec:hierarchical}

Building on our theoretical insights, we develop a hierarchical predictive coding framework that operates on the manifold of token embeddings to effectively capture the multi-scale structure of high-dimensional neural image data. Our framework is motivated by the principle of maximum information preservation across complementary representational dimensions. 

\textbf{Notation.} Let $\mathbf{x} = \{x_1, x_2, \ldots, x_K\}$ represent a sequence of visual tokens extracted from a volumetric EM image, where each $x_i \in \mathbb{R}^d$. For random token prediction, $\mathcal{M} \subset \{1, \ldots, K\}$ denotes the set of masked token indices with $|\mathcal{M}| = \rho K$ for masking ratio $\rho$, and $\mathcal{M}^c = \{1, \ldots, K\} \setminus \mathcal{M}$ denotes unmasked tokens. For sequential prediction, $x_{<i} = \{x_1, \ldots, x_{i-1}\}$ represents all tokens preceding position $i$.

We posit that optimal visual representation emerges from the integration of three complementary information channels, each capturing distinct aspects of the underlying neural structure. From an information-theoretic perspective, we can quantify the total extractable information as given in Eq.~\ref{eq:total_info}, where this formulation reveals how our three prediction tasks create a comprehensive representational space that captures position-invariant patterns, sequential dependencies, and global structural relationships simultaneously.

\subsubsection{Random Token Prediction}
At the microscopic level of visual structure, we employ stochastic token masking to induce the learning of position-invariant local feature detectors. Let $\mathcal{M} \subset \{1, \ldots, K\}$ denote a randomly sampled mask set with masking ratio $\rho$. The objective function maximizes the conditional log-likelihood of masked tokens given the unmasked context:
\begin{equation}
\mathcal{L}_{\text{random}} = -\mathbb{E}_{\mathcal{M} \sim \mathcal{D}_{\rho}} \left[\sum_{i \in \mathcal{M}} \log p_\theta(x_i | x_{\mathcal{M}^c})\right],
\end{equation}
where $\mathcal{D}_{\rho}$ is a distribution over mask patterns with expected density $\rho$.

This formulation encourages the development of a locally coherent latent space where proximal tokens exhibit statistical dependencies that reflect neuroanatomical regularities, creating a robust foundation for neuronal feature extraction by capturing repeating motifs in cellular membranes and organelles.

\subsubsection{Next Token Prediction}
To capture mesoscopic structural dependencies, we employ sequential autoregressive modeling along a predetermined tokenization path $\pi$. This objective maximizes:
\begin{equation}
\mathcal{L}_{\text{next}} = -\mathbb{E}_{\pi} \left[\sum_{i=1}^{K} \log p_\theta(x_{\pi(i)} | x_{\pi(<i)})\right],
\end{equation}
where $\pi(i)$ denotes the $i$-th token in the path, and $\pi(<i)$ represents all previous tokens in the sequence.

The path-conditional modeling captures critical transitional patterns in neuronal morphology, such as membrane continuity across adjacent regions and directional consistency in dendritic and axonal processes. This mesoscopic level of representation is particularly crucial for accurately tracking the anatomical continuity of neural structures across extended tissue volumes.

\subsubsection{Next-All Token Prediction}
To model macroscopic dependencies, we introduce a global prediction objective that forecasts all future tokens conditioned on the observed past:
\begin{equation}
\mathcal{L}_{\text{next-all}} = -\mathbb{E}_{\pi} \left[\sum_{i=1}^{K} \sum_{j=i}^{K} \log p_\theta(x_{\pi(j)} | x_{\pi(<i)})\right].
\end{equation}

This formulation induces the learning of latent variables that capture long-range correlations in neural architecture, crucial for modeling branching patterns, cell-type specific morphologies, and regionalized tissue organization. By integrating over all future positions, the model develops a holistic understanding of global structure rather than merely local transitions.

Importantly, this objective also provides a natural regularization against error accumulation in standard autoregressive models. For a sequence of length $K$, the expected prediction error under next-all modeling has a sublinear growth rate of $O(\sqrt{K})$ compared to the linear rate $O(K)$ in standard autoregressive approaches.

\subsubsection{Multi-Resolution Optimization Protocol}

To harmonize complementary objectives, we propose a principled multi-resolution optimization protocol following an easy-to-hard curriculum. We define a temporal modulation vector $\boldsymbol{\alpha}(t) = [\alpha(t), \beta(t), \gamma(t)] \in \mathbb{R}^3$ that governs task contributions at training iteration $t \in \mathbb{Z}^+$. After softmax normalization, each component satisfies $\alpha(t), \beta(t), \gamma(t) \in (0,1)$ and $\alpha(t) + \beta(t) + \gamma(t) = 1$.

The modulation vector evolves according to:
\begin{equation}
\boldsymbol{\alpha}(t) = \text{softmax}\left(\frac{\mathbf{v}(t)}{\tau(t)}\right),
\end{equation}
where $\mathbf{v}(t) = [v_\alpha(t), v_\beta(t), v_\gamma(t)] \in \mathbb{R}^3$ represents raw logits following a piecewise schedule, and $\tau(t) > 0$ is the annealing temperature controlling transition sharpness.

Our unified objective function is formulated as:
\begin{equation}
\mathcal{L}_{\text{TokenUnify}} = \alpha(t) \cdot \mathcal{L}_{\text{random}} + \beta(t) \cdot \mathcal{L}_{\text{next}} + \gamma(t) \cdot \mathcal{L}_{\text{next-all}}.
\end{equation}

The curriculum implements progressive transition from local to global representation learning through:
\begin{equation}
\mathbf{v}(t) = \begin{cases}
[2.0, -1.0, -2.0] & \text{if } t < T_1 \text{ (random-dominant)} \\
[-1.0, 2.0, -1.0] & \text{if } T_1 \leq t < T_2 \text{ (next-dominant)} \\
[-2.0, -1.0, 2.0] & \text{if } t \geq T_2 \text{ (next-all-dominant)}
\end{cases},
\end{equation}
where $T_1 = 0.3 \times T_{\text{total}}$ and $T_2 = 0.7 \times T_{\text{total}}$ are transition thresholds. With $\tau(t) = 1.0$, this yields dominant weights of $\sim$0.73 for the primary task while maintaining auxiliary contributions of $\sim$0.18 and $\sim$0.09 to preserve multi-task synergy.

For computational efficiency in next-all token prediction, we employ cross-attention with latent queries to aggregate sequence-wide information, enabling scalability to long token sequences without prohibitive complexity.

This hierarchical integration constructs a multi-resolution representational manifold capturing neural tissue organization across microscopic, mesoscopic, and macroscopic scales, providing robust foundations for volumetric electron microscopy circuit reconstruction.

\subsection{Finetuning and Segmentation}
\label{sec:finetune}
The segmentation network, denoted as $g_{\text{seg}}(\cdot)$, consists of an encoder $g_e(\cdot)$ and a decoder $g_d(\cdot)$:
\begin{equation}
g_{\text{seg}}(x; \theta_s) = g_d(g_e(x)), \quad \theta_s = \{\theta_e, \theta_d\},
\end{equation}
where $\theta_s$ represents the parameters of the entire segmentation network, and $\theta_e$ and $\theta_d$ are the parameters of the encoder and decoder, respectively.

The encoder $g_e(\cdot)$ gradually downsamples the input volume and extracts high-level semantic features, while the decoder $g_d(\cdot)$ upsamples the encoded features back to the original resolution. Meanwhile, the output of each downsampling layer in the encoder is connected to the corresponding layer in the decoder via skip connections to fuse local and global multi-scale information. 

We adopt 3D ResUNet/ViT/Mamba as the backbone network, respectively. For the Mamba variant, we adapt SegMamba~\cite{xing2024segmamba} for EM data, which we refer to as EMmamba, with enhanced decoder design for anisotropic EM features.

The output of the segmentation network $\hat{y} = g_{\text{seg}}(x) \in \mathbb{R}^{C \times D \times H \times W}$ represents the predicted affinity map~\cite{li2018contextual, li2023deception}, corresponding to the connectivity probability of each voxel in three directions. During training, the loss function for labeled samples is the mean squared error between the predicted and ground-truth affinity maps:
\begin{equation}
\mathcal{L}_{\text{seg}} = \frac{1}{|\mathcal{D}_l|}\sum_{i=1}^{|\mathcal{D}_l|} \|g_{\text{seg}}(x^l_i) - y_i\|^2.
\end{equation}
During inference, for any new test sample $x_t$, forward propagation through $g_{\text{seg}}(x_t)$ yields its predicted affinity map. This predicted affinity map is then post-processed using a seeded watershed algorithm and a structure-aware region agglomeration algorithm~\cite{beier2017multicut, funke2018large} to obtain the final neuron instance segmentation. Detailed information on our segmentation process and the segmentation pipeline are illustrated in the supplementary material.
\section{Datasets and Metrics}
\label{Data}


\begin{figure}[t]
    \centering
    \includegraphics[width=\linewidth]{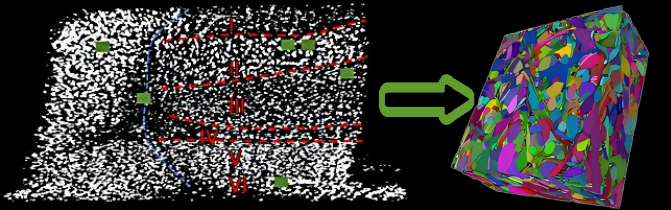}
    \put(-0.88\linewidth, -0.8em){(a) Sampling Area}
    \put(-0.33\linewidth, -0.8em){(b) Annotation}
    \caption{Illustration of the MEC dataset annotation. (a) shows the sampling locations of six different neural regions from the MEC raw data. Each region contains 1250 × 1250 ×125 voxels. (b) shows an example of the fine segmentation labels in one sampling region, which takes two experts one month to annotate.}
    \label{fig:dataset}
    \vspace{-0.3cm}
\end{figure}

\paragraph{Datasets.}
For pretraining TokenUnify, we collect over 1 TB of publicly available unlabeled EM imaging data from four large-scale datasets: FAFB \cite{schlegel2021automatic}, MitoEM \cite{wei2020mitoem}, FIB-25 \cite{takemura2017connectome}, and Kasthuri15 \cite{kasthuri2015saturated}. These datasets span diverse organisms including Drosophila, mouse, rat, and human samples, ensuring comprehensive coverage of EM morphology variations. We sample from these datasets with equal probability to guarantee diversity in pretraining data. Additional details are provided in supplementary material.

For downstream fine-tuning and evaluation, we employ two datasets: AC3/4 \cite{kasthuri2015saturated} and our large-scale MEC dataset. The AC3/4 dataset consists of 256/100 successive EM images with 1024×1024 spatial resolution from mouse somatosensory cortex. We partition AC4 into training (first 80 images) and validation (last 20 images), while using AC3 (first 100 images) for testing.

\begin{table*}[t]
\centering
{\fontsize{9}{11}\selectfont

\renewcommand{\arraystretch}{0.9}
\setlength{\tabcolsep}{1.8mm}
\begin{tabular}{r|l|cc>{\columncolor{Gray}}cc|cc>{\columncolor{Gray}}cc|c}
\toprule
\multirow{3}{*}{Post.} & \multirow{3}{*}{Method} & \multicolumn{4}{c|}{Wafer4} & \multicolumn{4}{c|}{Wafer36-2} & \multirow{3}{*}{\makecell{Param.\\(M)}}  \\
\cmidrule{3-10}
&  & {$VOI_M\downarrow$} & {$VOI_S\downarrow$} & \multirow{1}{*}{\makecell{$VOI\downarrow$}} & \multirow{1}{*}{$ARAND\downarrow$} & {$VOI_M\downarrow$} & {$VOI_S\downarrow$} & \multirow{1}{*}{\makecell{$VOI\downarrow$}} & \multirow{1}{*}{$ARAND\downarrow$} & \\
\midrule
& \multicolumn{10}{l}{\textbf{Supervised Methods}} \\
\cmidrule{2-11}
\multirow{5}{*}{\rotatebox{90}{Waterz \cite{funke2018large}}} 
& Superhuman \cite{lee2017superhuman} & 0.3328 & 1.1258 & 1.4587 & 0.1736 & 0.1506 & 0.4588 & 0.6094 & 0.0836 & 1.478 \\
& MALA \cite{funke2018large} & 0.5438 & 1.5027 & 2.0375 & 0.1115 & 0.3179 & 1.0664 & 1.3843 & 0.1570 & 84.03  \\
& PEA \cite{huang2022learning} & 0.3381 & 0.9276 & 1.2658 & 0.0677 & 0.2787 & 0.4279 & 0.7066 & 0.1169 & 1.480  \\
& UNETR \cite{hatamizadeh2022unetr} & 0.4504 & 1.6581 & 2.1085 & 0.2658 & 0.4478 & 0.5217 & 0.9696 & 0.2913 & 129.1 \\
& EMmamba & 0.4915 & 1.2924 & 1.7839 & 0.2052 & 0.2406 & 0.4189 & 0.6595 & 0.1231 & 28.30 \\
\cmidrule{2-11}
\multirow{5}{*}{\rotatebox{90}{LMC \cite{beier2017multicut}}}
& Superhuman \cite{lee2017superhuman} & 0.1948 & 1.9697 & 2.1644 & 0.2453 & 0.0792 & 1.1618 & 1.2410 & 0.1319 & 1.478 \\
& MALA \cite{funke2018large} & 0.3416 & 2.4129 & 2.7545 & 0.2567 & 0.1448 & 1.9603 & 2.1052 & 0.1977 & 84.03 \\
& PEA \cite{huang2022learning} & 0.1705 & 1.5993 & 1.7698 & 0.1527 & 0.4719 & 1.1226 & 1.5945 & 0.1588 & 1.480 \\
& UNETR \cite{hatamizadeh2022unetr} & 0.1791 & 3.1715 & 3.3506 & 0.6330 & 0.0949 & 1.3858 & 1.4807 & 0.1578 & 129.1 \\
& EMmamba & 0.1596 & 2.0580 & 2.2177 & 0.1973 & 0.0847 & 1.0351 & 1.1198 & 0.1253 & 28.30 \\
\midrule
& \multicolumn{10}{l}{\textbf{Self-Supervised Methods}} \\
\cmidrule{2-11}
\multirow{6}{*}{\rotatebox{90}{Waterz \cite{funke2018large}}}
& Random & 0.4915 & 1.2924 & 1.7839 & 0.2052 & 0.2406 & 0.4189 & 0.6595 & 0.1231 & \multirow{6}{*}{28.30} \\
& MAE \cite{he2022masked} & \underline{0.2325} & 1.0923 & 1.3248 & 0.0978 & 0.1629 & 0.4532 & 0.6161 & 0.0835 &  \\
& BYOL \cite{grill2020bootstrap} & 0.2584 & 0.9453 & 1.2037 & 0.0891 & 0.1748 & 0.4368 & 0.6116 & 0.0826 &  \\
& dbMIM \cite{chen2023self} & 0.2342 & \underline{0.8796} & \underline{1.1138} & \underline{0.0742} & \underline{0.1476} & \underline{0.3982} & \underline{0.5458} & \underline{0.0782} &  \\
& MS-Con-EM \cite{chen2024learning} & 0.2483 & 0.9102 & 1.1585 & 0.0813 & 0.1542 & 0.4037 & 0.5579 & 0.0794 &  \\
& TokenUnify & \textbf{0.1953} & \textbf{0.7998} & \textbf{0.9951} & \textbf{0.0509} & \textbf{0.1262} & \textbf{0.3585} & \textbf{0.4848} & \textbf{0.0650} &  \\
\cmidrule{2-11}
\multirow{6}{*}{\rotatebox{90}{LMC \cite{beier2017multicut}}}
& Random & 0.1596 & 2.0580 & 2.2177 & 0.1973 & \underline{0.0847} & \underline{1.0351} & \underline{1.1198} & 0.1253 & \multirow{6}{*}{28.30} \\
& MAE \cite{he2022masked} & \underline{0.1493} & 1.9216 & 2.0709 & 0.1583 & 0.0892 & 1.0927 & 1.1819 & 0.1243 &  \\
& BYOL \cite{grill2020bootstrap} & 0.1629 & 1.8374 & 2.0003 & 0.1427 & 0.0953 & 1.0782 & 1.1735 & 0.1287 &  \\
& dbMIM \cite{chen2023self} & 0.1573 & \underline{1.7436} & \underline{1.9009} & \underline{0.0913} & 0.0912 & 1.0649 & {1.1561} & \underline{0.1185} &  \\
& MS-Con-EM \cite{chen2024learning} & 0.1609 & 1.7692 & 1.9301 & 0.1026 & 0.0963 & 1.0703 & 1.1666 & 0.1207 &  \\
& TokenUnify & \textbf{0.1418} & \textbf{1.5103} & \textbf{1.6521} & \textbf{0.0591} & \textbf{0.0827} & \textbf{1.0276} & \textbf{1.1103} & \textbf{0.1074} &  \\
\bottomrule
\end{tabular}}
\caption{Quantitative comparison of electron microscopy segmentation results on Wafer4 and Wafer36-2 datasets. Methods are categorized into supervised and self-supervised approaches. All self-supervised methods use the same EMmamba backbone for fair comparison. ``Random'' refers to EMmamba without any pretraining. The best results are in \textbf{bold} and the second best results are \underline{underlined}.}
\label{tab:main1}
\end{table*}

To validate algorithm performance on large-scale EM data, we collected a 2TB MEC dataset by imaging mouse somatosensory cortex, medial entorhinal cortex, and cerebral cortex regions. We selected 6 representative volumes from different neural regions (wafer4/25/26/26-2/36/36-2), each with dimensions $1250 \times 1250 \times 125$ voxels. Dense manual annotation was performed on these regions, yielding over 1.2 billion annotated voxels through six months of work by two experts, as illustrated in Fig.~\ref{fig:dataset}. The dataset is split into training (wafer25/26/26-2/36), validation (wafer4), and testing (wafer36-2) sets.


\vspace{-0.4cm}
\paragraph{Metrics.}
To evaluate the performance of neuron segmentation \cite{zhang2024research,dang2024real}, we employ two widely-used metrics: Variation of Information (VOI) \cite{nunez2013machine} and Adjusted Rand Index (ARAND) \cite{arganda2015crowdsourcing}. These metrics quantify the agreement between the predicted segmentation and the ground truth from different perspectives. Detailed descriptions of these metrics can be found in the supplementary material.

\section{Experiment}
\label{sec:experiment}

We evaluate our proposed method on both MEC and AC3/4 datasets. Implementation details are provided in the supplementary material. We compare our method with both supervised and self-supervised approaches. For supervised methods, we include Superhuman \cite{lee2017superhuman}, MALA \cite{funke2018large}, PEA \cite{huang2022learning}, and UNETR \cite{hatamizadeh2022unetr}. 
For self-supervised methods, we evaluate two categories: (1) general-purpose visual representation learning methods including MAE \cite{he2022masked} and BYOL \cite{grill2020bootstrap}, which were originally designed for natural images but adapted to EM data; and (2) EM-specific self-supervised methods including dbMIM \cite{chen2023self} and MS-Con-EM \cite{chen2024learning}, which were specifically designed for electron microscopy images. All self-supervised methods utilize the same EMmamba backbone for fair comparison.

\begin{table}[t!]
\centering
\definecolor{Gray}{gray}{0.88}
\fontsize{8.3}{10}\selectfont
\renewcommand\tabcolsep{1.1pt}
\renewcommand{\arraystretch}{1.0}
\begin{tabular}{@{}l|cc>{\columncolor{Gray}}cc|c@{}}
\toprule
\multirow{1}{*}{Method} & {$VOI_M\downarrow$} & {$VOI_S\downarrow$} & \multirow{1}{*}{\makecell{$VOI\downarrow$}} & \multirow{1}{*}{$ARAND\downarrow$} & \multirow{1}{*}{\makecell{Param.(M)}} \\
\midrule
\multicolumn{6}{l}{\textbf{Supervised Methods}} \\
\midrule
Superhuman \cite{lee2017superhuman} & 0.4882 & 0.6563 & 1.1445 & 0.1748 & 1.478 \\
MALA \cite{funke2018large} & 0.4571 & 0.6767 & 1.1338 & 0.1664 & 84.02 \\
PEA \cite{huang2022learning} & 0.5522 & 0.4980 & 1.0502 & 0.2093 & 1.480 \\
UNETR \cite{hatamizadeh2022unetr} & 0.7799 & 0.7399 & 1.5198 & 0.2411 & 129.1 \\
EMmamba & 0.9378 & 0.8629 & 1.8007 & 0.2840 & 28.30 \\
\midrule
\multicolumn{6}{l}{\textbf{Self-Supervised Methods}} \\
\midrule
Random & 0.9378 & 0.8629 & 1.8007 & 0.2840 & \multirow{6}{*}{28.30} \\
MAE \cite{he2022masked} & \textbf{0.3167} & 0.7963 & 1.1131 & \textbf{0.1050} &  \\
BYOL \cite{grill2020bootstrap} & \underline{0.3842} & 0.7621 & 1.1463 & 0.1527 &  \\
dbMIM \cite{chen2023self} & 0.4623 & \underline{0.6279} & \underline{1.0902} & 0.1432 &  \\
MS-Con-EM \cite{chen2024learning} & 0.4785 & 0.6538 & 1.1323 & 0.1498 &  \\
TokenUnify & {0.4479} & \textbf{0.5439} & \textbf{0.9918} & \underline{0.1366} &  \\
\bottomrule
\end{tabular}
\caption{Quantitative comparison of segmentation results on AC3/4 datasets with waterz \cite{funke2018large} post-processing algorithm. All self-supervised methods use the same EMmamba backbone. ``Random'' refers to EMmamba without any pretraining. The best results are in \textbf{bold} and the second best results are \underline{underlined}.}
\label{tab:main2}
\end{table}

\begin{figure*}[t]
    \centering
    \includegraphics[width=\linewidth]{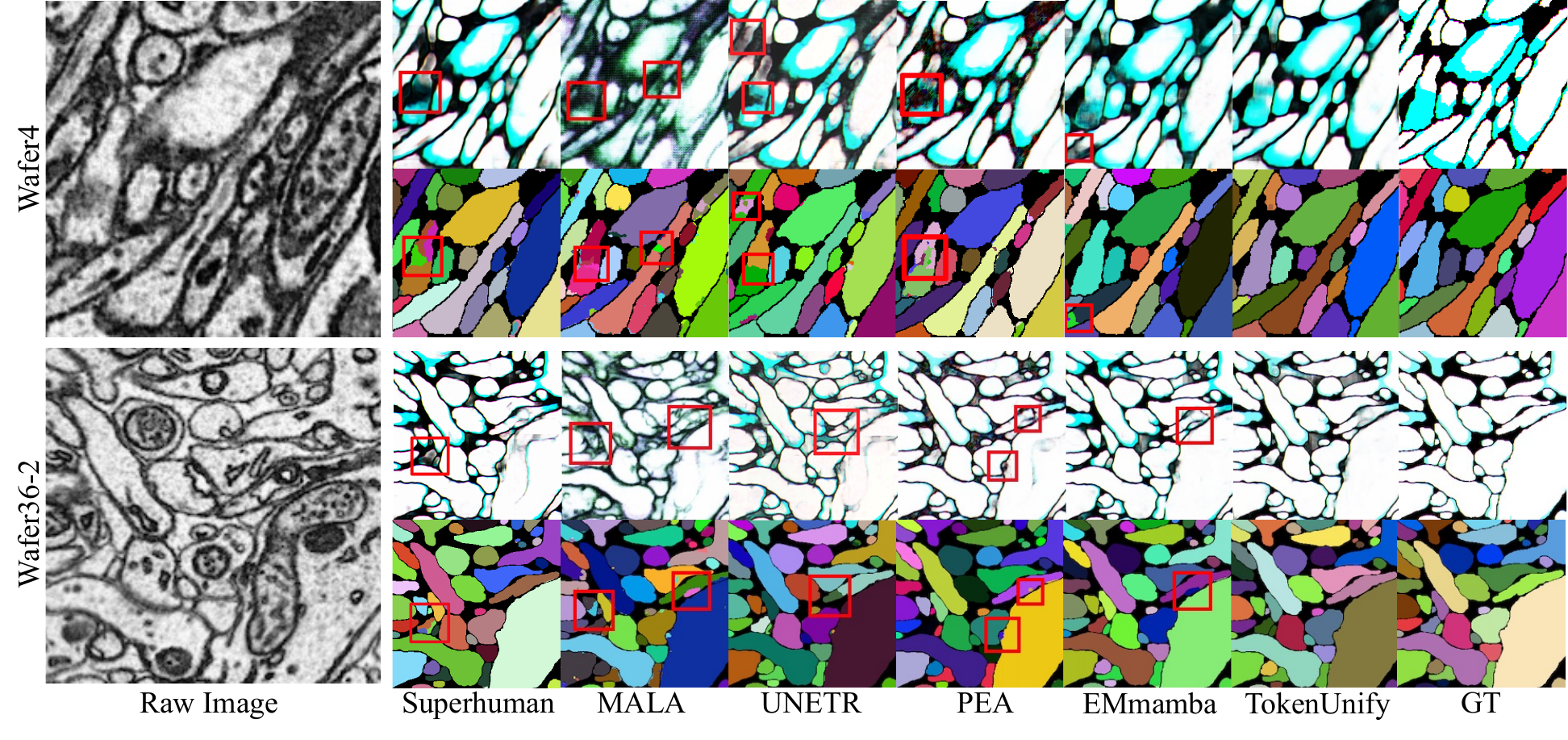}
    \caption{2D visualization results for two slices from the MEC dataset: wafer 4 and wafer 36-2. The left side displays the EM raw images, while the right side presents the affinity and segmentation results. \textcolor{red}{Red} boxes indicate the regions of mis-segmentation.}
    \vspace{-0.4cm}
    \label{fig:visual2d}
\end{figure*}

\begin{figure*}[t]
    \centering
    \includegraphics[width=\linewidth]{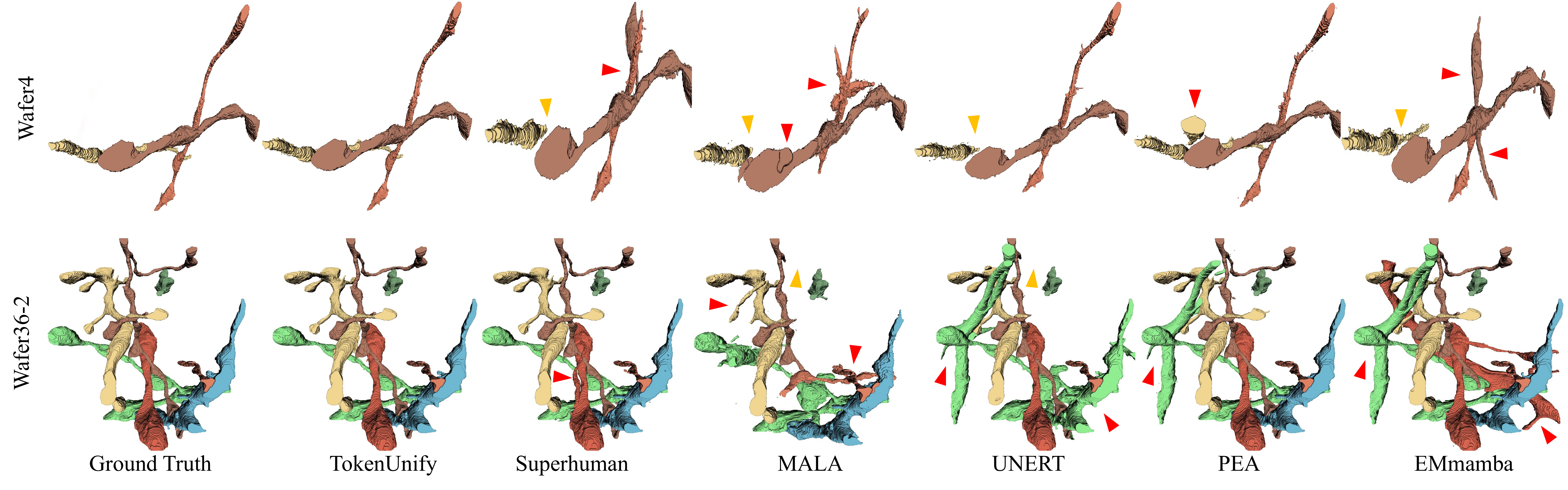}
    \caption{3D visualization results on two MEC datasets: wafer 4 and wafer 36-2. The \textcolor{red}{red} arrows indicate over-merged structures, and the \textcolor{orange}{orange} arrows highlight over-split structures.}
    \vspace{-0.4cm}
    \label{fig:visual3d}
\end{figure*}

\vspace{-0.4cm}
\paragraph{Results on MEC Dataset.} 
Table \ref{tab:main1} presents the quantitative results on the large-scale MEC dataset. Compared to models with similar parameter counts, on the Wafer4 volume under the waterz post-processing, our TokenUnify approach achieves approximately a 25\% performance improvement over MAE and a 44\% improvement compared to direct training without pretraining. Visualization results in Fig. \ref{fig:visual2d} and Fig. \ref{fig:visual3d} further demonstrate that our approach significantly outperforms other methods, particularly in challenging regions with complex neuronal structures.

\vspace{-0.4cm}
\paragraph{Results on AC3/4 Datasets.}
We also evaluate performance on the smaller AC3/4 dataset, which contains only about 1/10 of the labeled pixels compared to MEC. As shown in Table \ref{tab:main2}, even in this low-data scenario, the Mamba architecture combined with TokenUnify pretraining achieves performance comparable to state-of-the-art supervised methods like PEA. Additionally, it demonstrates a 11\% performance improvement over the MAE pretraining approach. This highlights the effectiveness of TokenUnify in scenarios with limited labeled data.

\begin{figure}[t]
    \centering
    \includegraphics[width = \linewidth]{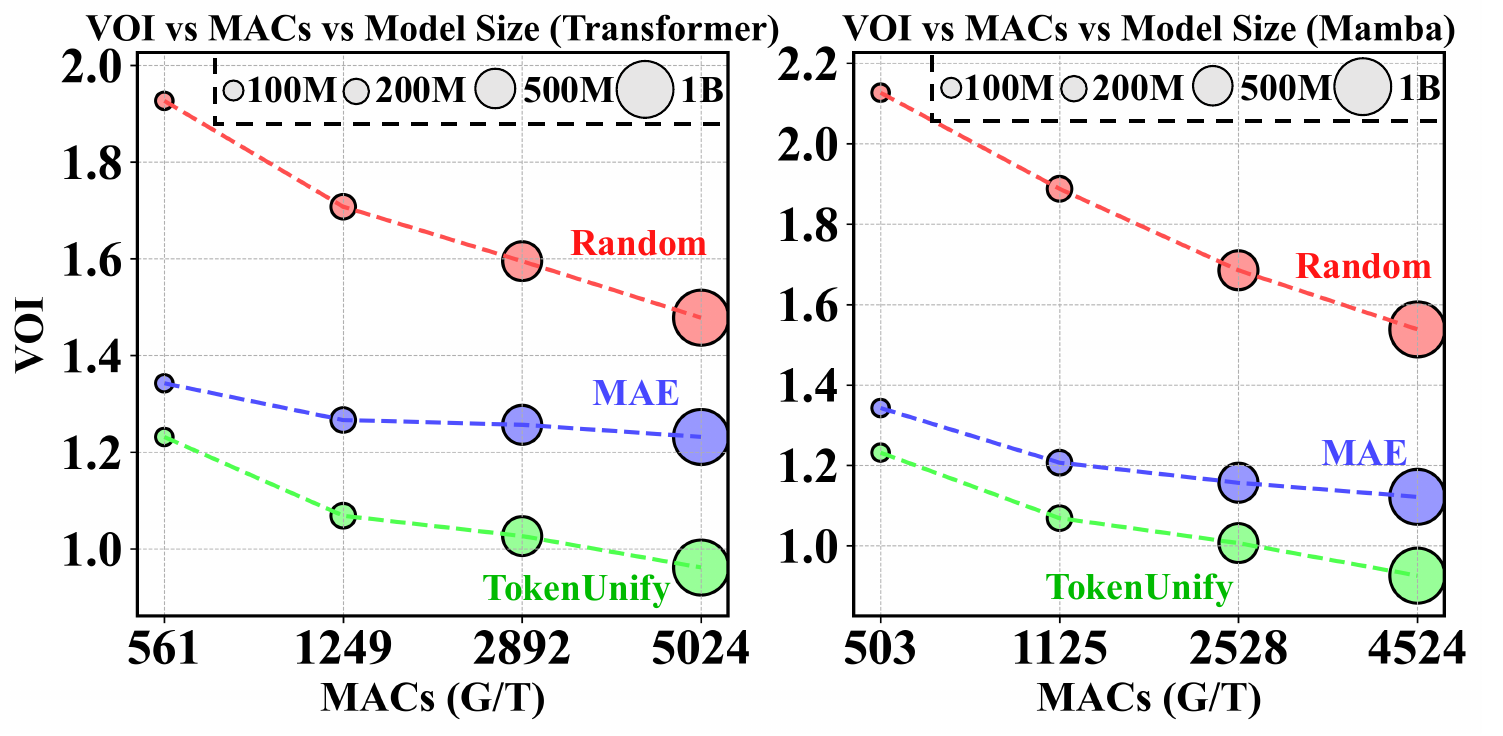}
    \vspace{-0.3cm}
    \caption{Performance evaluation of models with 100M, 200M, 500M, and 1B parameters. The models were trained for 100 epochs on the MEC dataset.}
    \vspace{-0.3cm}
    \label{fig:scalinglaw}
\end{figure}
\begin{table}[t]
\centering
\fontsize{8.5}{10}\selectfont
\renewcommand\tabcolsep{2pt}
\renewcommand{\arraystretch}{1}
\definecolor{Gray}{gray}{0.88}
\vspace{-0.3cm}
\begin{tabular}{ccccc}
\toprule
\multicolumn{3}{c}{Pretraining Strategy} & \multirow{2}{*}{$VOI\downarrow$} & \multirow{2}{*}{$ARAND\downarrow$}  \\
\cmidrule{1-3}
Random & Next & Next-all &  & \\
\midrule
\checkmark & & & 1.2680 & 0.0862 \\
 & \checkmark & & 4.0418 & 0.4416 \\
\checkmark & \checkmark & & \underline{1.1300} & \underline{0.0692} \\
\checkmark & & \checkmark & 1.1907 & 0.1203 \\
\rowcolor{Gray}\checkmark & \checkmark & \checkmark & \textbf{0.9951} & \textbf{0.0509} \\
\bottomrule
\end{tabular}
\caption{Ablation for different pretraining strategies on wafer4.}
\label{tab:ablation1}
\end{table}

\begin{table}[t]
\centering
\definecolor{Gray}{gray}{0.88}
\fontsize{8.5}{10}\selectfont
\renewcommand\tabcolsep{2pt}
\renewcommand{\arraystretch}{1}
\vspace{-0.3cm}
\begin{tabular}{ccccc}
\toprule
\multicolumn{3}{c}{Finetuning Module} & \multirow{2}{*}{$VOI\downarrow$} & \multirow{2}{*}{$ARAND\downarrow$}  \\
\cmidrule{1-3}
Mamba & Encoder & Decoder &  & \\
\midrule
\checkmark & & & 1.1362 & 0.0782 \\
 & \checkmark & & 1.5556 & 0.1370 \\
 & & \checkmark & 1.5295 & 0.1212 \\
\checkmark & \checkmark & & \underline{1.1065} & \underline{0.0629} \\
\rowcolor{Gray}\checkmark & \checkmark & \checkmark & \textbf{0.9951} & \textbf{0.0509} \\
\bottomrule
\end{tabular}
\caption{Ablation for the fine-tuning schemes on wafer4.}
\label{tab:ablation2}
\end{table}
\vspace{-0.4cm}
\paragraph{Scaling Law Analysis.}
To investigate how our method scales with model size, we evaluated various initialization and training strategies by scaling model parameters from 100M to 1B. Results in Fig. \ref{fig:scalinglaw} show that all methods exhibit performance improvements after pretraining. In contrast, TokenUnify demonstrates strong scaling properties, consistently outperforming other methods across both small and large model sizes. Notably, the Mamba architecture achieves competitive segmentation performance with fewer parameters compared to Transformer, confirming its efficiency for processing long sequence data in connectomics.

\subsection{Ablation Studies}
\label{sec11}

We conduct ablation experiments to analyze the key components of our approach. As shown in Table \ref{tab:ablation1} and Table \ref{tab:ablation2}, we examine two critical aspects:
\vspace{-0.4cm}
\paragraph{Ablation on Token Prediction Strategies.}
We systematically evaluate different token prediction combinations during pretraining on the wafer4 dataset (Table~\ref{tab:ablation1}). The full combination (Random + Next + Next-all) achieves optimal performance (VOI=0.9951), significantly outperforming individual strategies and demonstrating substantial improvements over baseline approaches. Next-token prediction alone performs poorly (VOI=4.0418), confirming that conventional autoregressive pretraining is insufficient for visual tasks. Unlike sequential text, visual data requires global spatial understanding that pure autoregressive modeling cannot capture effectively. Random token prediction provides better spatial initialization by encouraging spatial consistency, while the strategic combination of multiple prediction objectives yields more robust and generalizable representations for segmentation. These results validate our hypothesis that diverse pretraining tasks are essential for effective visual representation learning.
\vspace{-0.4cm}
\paragraph{Fine-tuning Components.}  
We also investigate which network components benefit most from fine-tuning (Table~\ref{tab:ablation2}). Our architecture consists of three main parts: Mamba blocks (sequence modeling), encoder (downsampling), and decoder (upsampling). 
Fine-tuning only the Mamba blocks yields substantially better results (VOI=1.1362) than fine-tuning only the encoder (VOI=1.5556) or decoder (VOI=1.5295), demonstrating the critical role of sequence modeling in connectomics segmentation. For resource-constrained scenarios, prioritizing the fine-tuning of Mamba blocks provides an efficient balance between performance and computational cost.

\section{Conclusion}
We propose TokenUnify, a framework that unifies random token prediction, next-token prediction, and next-all token prediction to capture multi-scale visual structures. Our analysis shows this approach mitigates cumulative errors in autoregressive modeling. To validate TokenUnify, we introduce a large-scale EM neuron segmentation dataset with 1.2 billion labeled voxels across six mouse brain regions. Experiments demonstrate a 44\% improvement over training from scratch and a 25\% gain over MAE pretraining. Scaling analysis further highlights its efficiency with increasing model size and data volume. By releasing our dataset and implementation, we aim to advance biological imaging and inspire future research in visual representation learning.

\section*{Acknowledgments}
This work was supported by the National Natural Science Foundation of China under Grant 624B2137. We thank the staff at the Institute of Artificial Intelligence, Hefei Comprehensive National Science Center for their assistance with biological specimen preparation, imaging, and annotation. We are grateful to the anonymous reviewers for their thorough reviews and constructive suggestions.

{
    \small
    \bibliographystyle{ieeenat_fullname}
    \bibliography{main}
}
\clearpage  
\onecolumn  
\clearpage
\setcounter{page}{1}
\maketitlesupplementary

\tableofcontents

\section{Detailed Information about Datasets and Metrics}
\subsection{Datasets}
\begin{figure}[t]
    \centering
    \includegraphics[width=\linewidth]{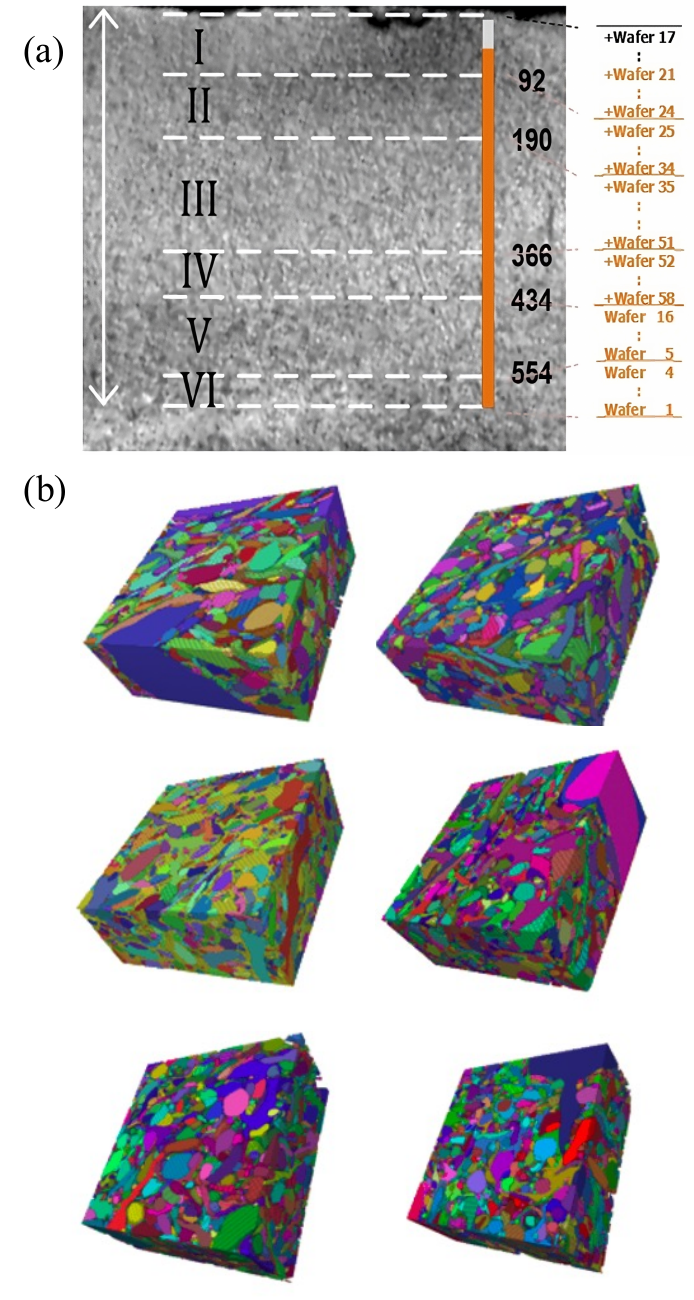}
    \caption{The relative positions of the wafer layers selected from the MEC dataset.}
    \label{fig:MEC}
\end{figure}

\subsubsection{Pretraining Data Organization}
This chapter serves as a supplement to Section 5 in the main paper, providing detailed information about the datasets used in this study. 

For the pretraining phase of TokenUnify, we additionally leverage a diverse collection of publicly available unlabeled EM imaging data from four large-scale EM datasets: FAFB~\cite{schlegel2021automatic}, MitoEM~\cite{wei2020mitoem}, FIB-25~\cite{takemura2017connectome}, and Kasthuri15~\cite{kasthuri2015saturated}. These datasets cover a wide range of organisms, including Drosophila, mouse, rat, and human samples, totaling over 1 TB of high-resolution EM data. The diversity of this pretraining data ensures that our model learns robust features that generalize across different brain regions and even different species.

We sample from these datasets with equal probability during pretraining, guaranteeing the diversity of the visual features encountered by the model. This comprehensive pretraining strategy enables TokenUnify to learn generalizable representations of neuronal structures that can be effectively fine-tuned for specific segmentation tasks.

All pretraining datasets employed are publicly available, with their specifics outlined in Table \ref{tab:EM2}.
\begin{table*}[h]
\centering
\renewcommand\tabcolsep{3.2pt}
\renewcommand{\arraystretch}{1.2}
\begin{tabular}{l c c c c}
\toprule[1.2pt]
Dataset & Modality & Resolution & Species & Target Region \\ \midrule
Full Adult Fly Brain (FAFB) \cite{schlegel2021automatic} & EM & 4 \(\times\) 4 \(\times\) 40 \(nm^3\) & \textit{Drosophila} & Whole brain \\
MitoEM-H \cite{wei2020mitoem} & EM & 8 \(\times\) 8 \(\times\) 30 \(nm^3\) & Human & Cortex (Mitochondria) \\
MitoEM-R \cite{wei2020mitoem} & EM & 8 \(\times\) 8 \(\times\) 30 \(nm^3\) & Rat & Cortex (Mitochondria) \\
FIB-25 \cite{takemura2017connectome} & EM & 5 \(\times\) 5 \(\times\) 5 \(nm^3\) & \textit{Drosophila} & CA1 Hippocampus \\
Kasthuri15 \cite{kasthuri2015saturated} & EM & 3 \(\times\) 3 \(\times\) 30 \(nm^3\) & Mouse & Neocortex \\
\bottomrule[1.2pt]
\end{tabular}
\caption{Detailed description of the EM pre-taining datasets}
\label{tab:EM2}
\end{table*}

\subsection{Ultra-high Resolution EM Dataset MEC Construction}
To support the development and evaluation of our hierarchical predictive coding framework, we introduce a large-scale electron microscopy (EM) dataset specifically designed to capture the long-range spatial dependencies critical for neuron segmentation. The construction of this dataset addresses a fundamental challenge in the field: the lack of comprehensive, finely annotated EM data with sufficient scale to train and evaluate models that can capture complex neuronal structures.
\subsubsection{Data Collection}
The MEC dataset originates from our team's Mouse MEC MultiBeam-SEM imaging efforts, where we performed comprehensive brain imaging of mice, accumulating data at the petabyte scale.
MEC dataset consists of high-resolution EM images acquired from multiple regions of the mouse brain. Using advanced sample preparation techniques and state-of-the-art electron microscopy, we collected a 2TB dataset imaging the mouse somatosensory cortex, mouse medial entorhinal cortex, and mouse cerebral cortex at a resolution of 4nm×4nm×35nm per voxel. This ultra-high resolution enables the visualization of fine neuronal structures, including dendritic spines, axonal boutons, and synaptic connections that are essential for understanding neural circuits.

\subsubsection{Large-Scale Manual Annotation}
To provide ground truth for training and evaluation, we conducted extensive manual annotation of the EM volumes. As shown in Fig.~\ref{fig:MEC}(b), we selected six representative volumes from different neural regions, named wafer4/25/26/26-2/36/36-2 as illustrated in Fig. \ref{fig:MEC}(a), with each volume size reaching 1250 × 1250 × 125 voxels. These regions were carefully chosen to represent diverse neuronal morphologies and circuit organizations, ensuring that models trained on this data can generalize to various brain structures.

The annotation process involved precise delineation of neuronal boundaries by expert neuroscientists, identifying distinct neurons as separate instances while preserving their complex morphological features. This labor-intensive process took two experts a total of six months to complete, resulting in over 1.2 billion annotated voxels. The annotation pipeline involved multiple quality control steps to ensure consistency and accuracy, including cross-validation between annotators and verification against known neuroanatomical structures.

\subsubsection{Spatial Continuity for Long-sequence Modeling}
A key feature of our MEC dataset is its emphasis on spatial continuity, making it an ideal testbed for evaluating methods that aim to capture long-range dependencies. Unlike many existing computer vision datasets that consist of independent images, our EM volumes preserve the natural continuity of neuronal structures across thousands of consecutive slices. This continuity is essential for modeling the complex branching patterns and long-range connections characteristic of neuronal morphology.

The ultra-high resolution of our dataset allows for the extraction of thousands of continuous image tokens from a single volume, providing the necessary context length to evaluate autoregressive models. This property makes our dataset particularly well-suited for TokenUnify, which is designed to leverage both local and global context in predicting complex visual structures.


\subsection{Metrics}
\label{sec:metrci}
Variation of Information (VOI) is an information-theoretic measure that assesses the distance between two clusterings in terms of their average conditional entropy. Given the predicted segmentation $S_{pred}$ and the ground-truth segmentation $S_{gt}$, VOI is defined as:
\begin{equation}
VOI(S_{pred}, S_{gt}) = H(S_{pred}|S_{gt}) + H(S_{gt}|S_{pred}),
\end{equation}
where $H(\cdot|\cdot)$ denotes the conditional entropy. It can be calculated by:
\begin{multline}
H(S_{pred}|S_{gt}) = \\ - \sum_{i=1}^{|S_{gt}|} \sum_{j=1}^{|S_{pred}|} \frac{|S_{gt}^i \cap S_{pred}^j|}{N} \log \frac{|S_{gt}^i \cap S_{pred}^j|}{|S_{gt}^i|},
\end{multline}
where $S_{gt}^i$ and $S_{pred}^j$ represent the $i$-th and $j$-th segments in the ground-truth and predicted segmentation, respectively, and $N$ is the total number of voxels. VOI ranges from 0 to $\infty$, with a lower value indicating better segmentation performance.

Adjusted Rand Index (ARAND) is a variant of the Rand Index \cite{arganda2015crowdsourcing} that corrects for chance when comparing two clusterings. It is defined as:
\begin{multline}
ARAND(S_{pred}, S_{gt}) = \\ \frac{\sum_{ij} \binom{n_{ij}}{2} - [\sum_i \binom{a_i}{2} \sum_j \binom{b_j}{2}] / \binom{N}{2}}{[\sum_i \binom{a_i}{2} + \sum_j \binom{b_j}{2}] / 2 - [\sum_i \binom{a_i}{2} \sum_j \binom{b_j}{2}] / \binom{N}{2}},
\end{multline}
where $n_{ij}$ is the number of voxels that are in segment $i$ of $S_{pred}$ and segment $j$ of $S_{gt}$, $a_i = \sum_j n_{ij}$ is the number of voxels in segment $i$ of $S_{pred}$, $b_j = \sum_i n_{ij}$ is the number of voxels in segment $j$ of $S_{gt}$, and $N = \sum_{ij} n_{ij}$ is the total number of voxels. ARAND ranges from 0 to 1, with a lower value indicating better segmentation performance.

\begin{figure*}[t]
    \centering
    \includegraphics[width = \linewidth]{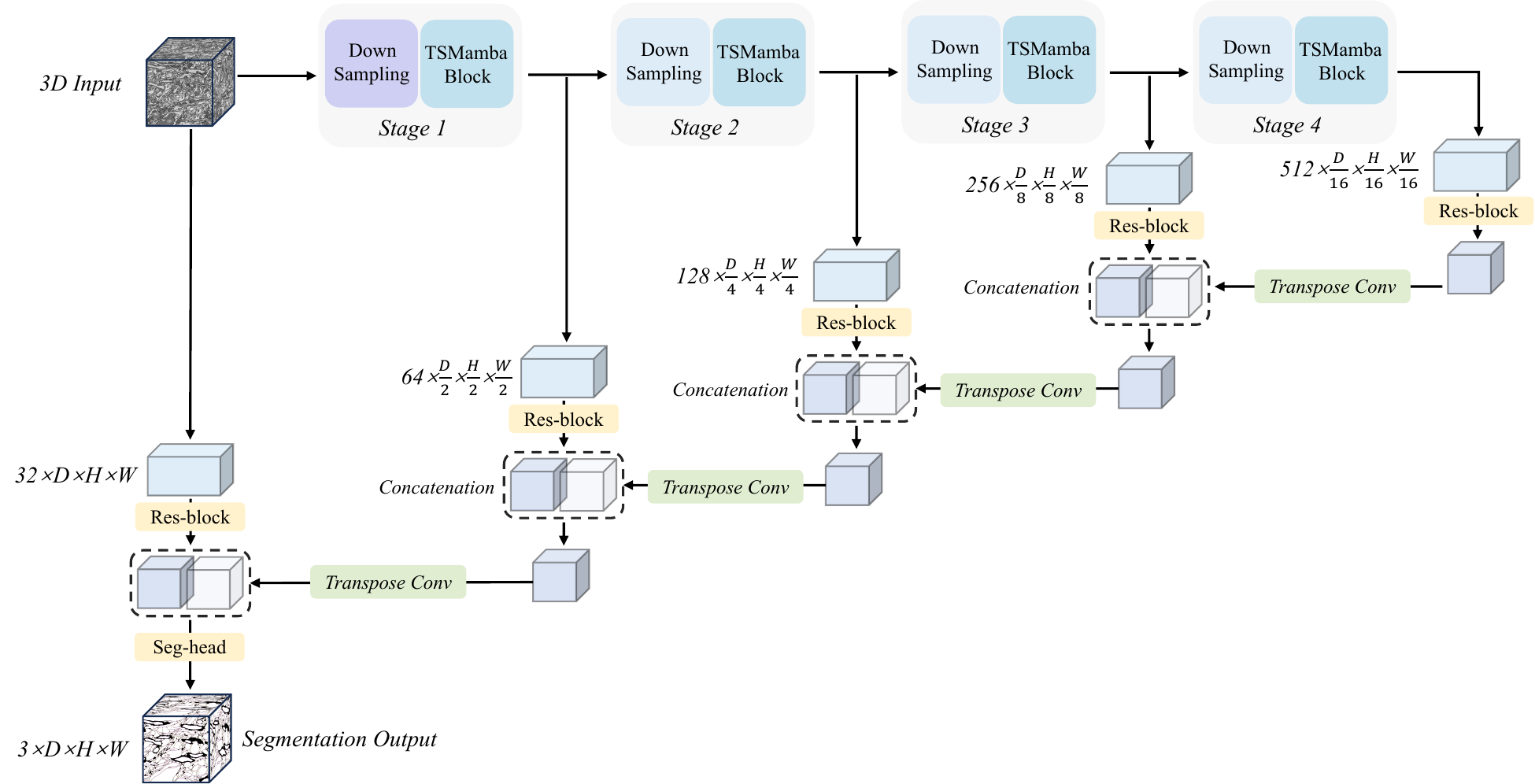}
    \caption{Segmentation pipeline.}
    \label{fig:segpipe}
\end{figure*}

\begin{table*}[t]
\centering
\renewcommand\tabcolsep{2.8pt}
\renewcommand{\arraystretch}{1.3}
\begin{tabular}{lccccc}
\toprule[1.2pt]
& EMmamba-tiny & EMmamba-small & EMmamba-middle & EMmamba-large & EMmamba-huge \\
\midrule
Mamba layer & {[}2,2,2,2{]} & {[}2,2,2,2{]} & {[}2,2,2,2{]} & {[}2,2,2,2{]} & {[}2,2,2,2{]} \\
Feature size & {[}32,64,128,256{]} & {[}64,128,256,512{]} & {[}96,192,384,768{]} & {[}144,288,576,1104{]} & {[}192,384,768,1536{]} \\
Hidden size & 512 & 1024 & 1024 & 2048 & 3072 \\
Kernel size  & {[}1,5,5{]} & {[}1,5,5{]} & {[}1,5,5{]} & {[}1,5,5{]} & {[}1,5,5{]} \\
Batch size & 40 & 22 & 12 & 8  & 4  \\
Param. (M)& 28.30 & 112.5 & 206.6 & 506.6 & 1008 \\
\bottomrule[1.2pt]
\end{tabular}
\caption{Shows the differ in architecture when adding the parameters of the segmentation backbone.}
\label{tab: modelsize}
\end{table*}

\section{Method Details}
\paragraph{Implementation Details.}
\label{details}
We employ consistent training configurations for both pretraining and fine-tuning phases. The network architecture remains unchanged throughout all training stages. For fine-tuning, we optimize using AdamW optimizer~\cite{loshchilov2018decoupled} with $\beta_1=0.9$, $\beta_2=0.999$, learning rate of $1 \times 10^{-6}$, and batch size of 20 on NVIDIA GTX 3090 (24GB) GPUs. Pretraining utilizes batch size of 8 on NVIDIA Tesla A40 (48GB) GPUs due to memory constraints.

We conduct distributed training with 8 NVIDIA GTX 3090 GPUs for segmentation tasks (1200 epochs) and 32 NVIDIA Tesla A40 GPUs for pretraining tasks (400 epochs). The pretraining input volume resolution is set to $16 \times 160 \times 160$ voxels with patch size of $4 \times 16 \times 16$ voxels for tokenization.

\textbf{Multi-Resolution Optimization Protocol.} Our hierarchical predictive coding employs a temporal modulation strategy with task weights $\boldsymbol{\alpha}(t) = [\alpha(t), \beta(t), \gamma(t)]$ governing the contributions of random token prediction, next-token prediction, and next-all token prediction respectively. The curriculum follows an easy-to-hard progression:
\begin{align}
\boldsymbol{\alpha}(t) = \begin{cases}
[0.73, 0.18, 0.09] & \text{if } t < T_1 \text{ (random-dominant)} \\
[0.18, 0.73, 0.09] & \text{if } T_1 \leq t < T_2 \text{ (next-dominant)} \\
[0.09, 0.18, 0.73] & \text{if } t \geq T_2 \text{ (next-all-dominant)}
\end{cases}
\end{align}
where transition thresholds are $T_1 = 0.3 \times T_{\text{total}}$ and $T_2 = 0.7 \times T_{\text{total}}$. This progressive weighting scheme implements our theoretical motivation that local feature learning should precede global structure modeling.

For downstream segmentation, we employ two post-processing algorithms: Waterz~\cite{funke2018large} with 50\% quantile threshold and LMC~\cite{beier2017multicut} using Kernighan-Lin solver~\cite{kernighan1970efficient}. Network initialization for fine-tuning loads pretrained weights following established protocols~\cite{he2022masked}.
\begin{algorithm}[h]
\caption{TokenUnify Pre-training}
\label{pretrainal}
\SetKwInOut{Input}{Input}
\SetKwInOut{Output}{Output}

\Input{Unlabeled image data $X = \{X^{(1)}, \dots, X^{(T)}\}$}
\Input{Model parameters $\theta_1$}
\Output{Pre-trained model $f_{\theta_1}(\cdot)$}

\BlankLine
\For{$t \gets 1$ \KwTo $T$}{
    Partition $X^{(t)}$ into patches $\{x_1, \dots, x_K\}$ \\
    Tokenize patches: $\{x_1, \dots, x_K\} \rightarrow \text{tokens}$ \\
    \BlankLine
    \textbf{Compute loss functions:} \\
    Random token prediction: $\mathcal{L}_{\text{random}} = -\sum_{i \in M} \log p(x_i \mid x_{\bar{M}})$ \\
    Next token prediction: $\mathcal{L}_{\text{next}} = -\sum_{i=1}^K \log p(x_i \mid x_{<i})$ \\
    Next-all token prediction: $\mathcal{L}_{\text{next-all}} = -\sum_{i=1}^K \sum_{j=i}^K \log p(x_j \mid x_{<i})$ \\
    \BlankLine
    Update $\theta_1$ to minimize $\mathcal{L}_{\text{random}}$, $\mathcal{L}_{\text{next}}$, $\mathcal{L}_{\text{next-all}}$
}

\Return $f_{\theta_1}(\cdot)$
\end{algorithm}

Pre-training is conducted on a large-scale, ultra-high-resolution electron microscopy (EM) image dataset, providing spatially correlated long sequences. TokenUnify demonstrates significant improvements in segmentation performance on downstream EM neuron segmentation tasks compared to existing methods. Our pre-training and fine-tuning algorithms are summarized in Algorithm \ref{pretrainal} and Algorithm \ref{finetuneal}, respectively.
\begin{algorithm}[h]
\caption{TokenUnify Fine-tuning}
\label{finetuneal}
\SetKwInOut{Input}{Input}
\SetKwInOut{Output}{Output}

\Input{Labeled data $\mathcal{D}_l = \{(x_i^l, y_i)\}_{i=1}^{|\mathcal{D}_l|}$}
\Input{Pre-trained model $f_{\theta_1}(\cdot)$}
\Input{Segmentation model $g_{\theta_2}(\cdot)$}
\Output{Fine-tuned segmentation model $g_{\theta_2}(\cdot)$}

\BlankLine
Initialize $\theta_2$ with $\theta_1$ \\
\BlankLine
\For{$i \gets 1$ \KwTo $|\mathcal{D}_l|$}{
    $\hat{y}_i = g_{\theta_2}(f_{\theta_1}(x_i^l))$ \\
    $\mathcal{L}_{\text{seg}} = \frac{1}{|\mathcal{D}_l|} \sum_{i=1}^{|\mathcal{D}_l|} |\hat{y}_i - y_i|^2$ \\
    Update $\theta_2$ to minimize $\mathcal{L}_{\text{seg}}$
}

\Return $g_{\theta_2}(\cdot)$
\end{algorithm}
The TokenUnify pre-training algorithm captures both local and global dependencies in image data through mixed token prediction tasks. The Mamba network architecture ensures efficient modeling of long sequences. During fine-tuning, the pre-trained model adapts to downstream segmentation tasks using labeled data, achieving state-of-the-art performance on EM neuron segmentation benchmarks.
\subsection{Perceiver Resampler}
\label{sec:pr}
The workflow of the Perceiver Resampler \cite{alayrac2022flamingo,chen2024automated,chen2023quantifying} can be summarized in the following steps:
1. Combine the output of the Vision Encoder (e.g., features from images) with learned time position encodings.
2. Flatten the combined features into a one-dimensional sequence.
3. Process the flattened features using Transformer layers that incorporate attention mechanisms, which interact with learned latent query vectors.
Output a fixed number of visual tokens equal to the number of latent queries.

\begin{algorithm}[h]
\caption{Perceiver Resampler Pseudocode}
\SetKwInOut{Input}{Input}
\SetKwInOut{Output}{Output}

\Input{$\mathbf{x}_f$ - The [T, S, d] visual features (T=time, S=space)}
\Input{$\mathbf{t}$ - The [T, 1, d] time position embeddings}
\Input{$\mathbf{x}$ - R learned latents of shape [R, d]}
\Input{$\text{num\_layers}$ - Number of layers}
\Output{$\mathbf{x}$ - Updated learned latents}

\BlankLine
\textbf{Add time position embeddings and flatten:} \\
$\mathbf{x}_f \gets \mathbf{x}_f + \mathbf{t}$ \\
$\mathbf{x}_f \gets \text{flatten}(\mathbf{x}_f)$ \\
\ \ \ \ \tcp*[h]{\([T, S, d] \rightarrow [T \times S, d]\)}

\BlankLine
\textbf{Apply the Perceiver Resampler layers:} \\
\For{$i \gets 1$ \KwTo $\text{num\_layers}$}{
    $\mathbf{x} \gets \mathbf{x} + \text{attention}_i(q=\mathbf{x}, kv=\text{concat}([\mathbf{x}_f, \mathbf{x}]))$ \\
    $\mathbf{x} \gets \mathbf{x} + \text{ffw}_i(\mathbf{x})$
}

\Return $\mathbf{x}$
\end{algorithm}
The input visual features, denoted as $\mathbf{x}_f$, have a shape of \([T, S, d]\), where \(T\) represents the time dimension, \(S\) the spatial dimension, and \(d\) the feature dimension. The time position embeddings, represented by $\mathbf{t}$, are of shape \([T, 1, d]\) and are added to the visual features to incorporate temporal information.

The learned latents, denoted as $\mathbf{x}$, have a shape of \([R, d]\), where \(R\) is the number of latents and \(d\) is the feature dimension. The parameter \texttt{num\_layers} specifies the number of layers in the Perceiver Resampler model.

The operation \texttt{flatten} reshapes the input tensor from \([T, S, d]\) to \([T \times S, d]\). The function \texttt{attention\_i} represents the attention mechanism applied in the \(i\)-th layer, which takes a query \(q\) and key-value pairs \(kv\). The function \texttt{concat} concatenates the input tensors along the specified dimension. Finally, \texttt{ffw\_i} refers to the feedforward network applied in the \(i\)-th layer.
\subsection{Alternating Protocol}
Our alternating protocol implements stochastic mode switching:

\begin{algorithm}[t]
\SetAlgoNoLine
\caption{Alternating Optimization Protocol}
\label{alg:alternating} 
\SetKwInOut{Input}{Input}
\SetKwProg{Fn}{Function}{}{}

\Input{Training data $\mathcal{D}$, model $f_\theta$, initial weights $\theta_0$}
\BlankLine
Pretrain using MAE phases 1-2 \\

\For{$t = 1$ \KwTo $T$}{
    Sample mode $m_t \sim P_t$ where $P_t = [p_{\text{next-all}}, p_{\text{AR}}, p_{\text{Mask}}]^\top$ \\
    Compute batch loss $\mathcal{L}_{m_t}$ for current mode \\
    Update $\theta_{t+1} \gets \theta_t - \eta_t \nabla_\theta \mathcal{L}_{m_t}$ \\
    Anneal $P_t$: Increase $p_{\text{AR}}$ while decreasing $p_{\text{Mask}}$ \tcp*{Probability adjustment}
}
\end{algorithm}

The sampling distribution $P_t$ follows a curriculum schedule:
\begin{equation}
    p_{\text{Mask}} = \max(0.7 - t/\tau, 0.3), \quad p_{\text{AR}} = 1 - p_{\text{Mask}} - p_{\text{next-all}}
\end{equation}
where $\tau$ is the transition period hyperparameter. This implements gradual shift from reconstruction-heavy to prediction-focused training.

\begin{theorem}[Convergence Guarantee]
\label{thm:convergence}
Let $\mathcal{L}_t$ satisfy $\|\nabla\mathcal{L}_t - \nabla\mathcal{L}_{t+1}\| \leq L\|\theta_t - \theta_{t+1}\|$ with step sizes $\eta_t = \eta_0/\sqrt{t}$. Then alternating optimization achieves:
\begin{equation}
    \min_{1 \leq t \leq T} \mathbb{E}[\|\nabla\mathcal{L}_t\|_2] \leq \frac{C}{\sqrt{T}} \left(1 + \log T + \sigma^2_{\text{mode}}\right)
\end{equation}
where $C$ is a constant and $\sigma^2_{\text{mode}}$ quantifies mode sampling variance.
\end{theorem}

Proof sketch appears in Appendix~\ref{thm:formal_convergence}, extending \cite{nesterov2018lectures} to our alternating regime. The bound shows sublinear convergence despite mode switching stochasticity.

\subsection{Segmentation Method}
\label{segsection}
The EMmamba network is structured into three principal components (as detailed in Fig. \ref{fig:segpipe}): 3D feature encoder, convolution-based decoder for segmentation prediction, and skip connections to integrate local multi-scale features into the decoder for feature fusion \cite{liu2023deep,liu2023toothsegnet,sun2024eagle}. 

To achieve effective feature encoding, we designed anisotropic downsampling layers and adopted the TSMamba block from the Segmamba \cite{xing2024segmamba}. Specifically, in Stage 1, the downsampling layer uses a convolutional kernel size of (1, 7, 7). For the subsequent three layers, the downsampling layers have a convolutional kernel size of (1, 2, 2). The decoder section employs a convolutional kernel size of (1, 5, 5). This anisotropic design is particularly advantageous for processing EM images, which exhibit inherent anisotropy. And the detailed network structures of different parameters are provided in Table \ref{tab: modelsize}.

\begin{table*}[t]
\centering
{\fontsize{9}{11}\selectfont
\definecolor{Gray}{gray}{0.88}
\renewcommand{\arraystretch}{0.9}

{
\begin{tabular}{@{}r|l|cc>{\columncolor{Gray}}cc@{}}    
\toprule
\multirow{3}{*}{Post.} & \multirow{3}{*}{Method} & \multicolumn{4}{c}{Wafer4} \\
\cmidrule{3-6}
&  & {$VOI_M\downarrow$} & {$VOI_S\downarrow$} & \multirow{1}{*}{\makecell{$VOI\downarrow$}} & \multirow{1}{*}{$ARAND\downarrow$} \\
\midrule
& \multicolumn{5}{l}{\textbf{ Supervised Methods}} \\    
\cmidrule{2-6} 
\multirow{5}{*}{\rotatebox{90}{Waterz \cite{funke2018large}}} 
& Superhuman \cite{lee2017superhuman} & 0.3392$\pm$0.0167 & 1.2247$\pm$0.0857 & 1.5639$\pm$0.0921 & 0.2050$\pm$0.0284 \\
& MALA \cite{funke2018large} & 0.6217$\pm$0.1266 & 1.5314$\pm$0.1123 & 2.1531$\pm$0.1004 & 0.1490$\pm$0.0476 \\
& PEA \cite{huang2022learning} & 0.3943$\pm$0.0655 & 1.0036$\pm$0.1435 & 1.3979$\pm$0.2090 & 0.0963$\pm$0.0310 \\
& UNETR \cite{hatamizadeh2022unetr} & 0.4454$\pm$0.0155 & 1.7979$\pm$0.1548 & 2.2433$\pm$0.1424 & 0.3244$\pm$0.0701 \\
& EMmamba & 0.4353$\pm$0.0520 & 1.3018$\pm$0.0086 & 1.7371$\pm$0.0432 & 0.1872$\pm$0.0156 \\
\cmidrule{2-6}    
\multirow{5}{*}{\rotatebox{90}{LMC \cite{beier2017multicut}}}
& Superhuman \cite{lee2017superhuman} & 0.2006$\pm$0.0054 & 2.1283$\pm$0.1378 & 2.3289$\pm$0.1427 & 0.2924$\pm$0.0408 \\
& MALA \cite{funke2018large} & 0.3094$\pm$0.0478 & 2.3802$\pm$0.1863 & 2.6869$\pm$0.1558 & 0.2303$\pm$0.0314 \\
& PEA \cite{huang2022learning} & 0.2303$\pm$0.0870 & 1.6373$\pm$0.1289 & 1.8343$\pm$0.0732 & 0.1611$\pm$0.0152 \\
& UNETR \cite{hatamizadeh2022unetr} & 0.1625$\pm$0.0144 & 3.3146$\pm$0.1391 & 3.4772$\pm$0.1272 & 0.6600$\pm$0.0304 \\
& EMmamba & 0.1594$\pm$0.0005 & 2.0921$\pm$0.0300 & 2.2515$\pm$0.0298 & 0.2104$\pm$0.0113 \\
\midrule
& \multicolumn{5}{l}{\textbf{ Self-Supervised Methods}} \\    
\cmidrule{2-6}    
\multirow{6}{*}{\rotatebox{90}{Waterz \cite{funke2018large}}}
& Random & 0.4353$\pm$0.0520 & 1.3018$\pm$0.0086 & 1.7371$\pm$0.0432 & 0.1872$\pm$0.0156 \\
& MAE \cite{he2022masked} & 0.2363$\pm$0.0212 & 1.0782$\pm$0.0251 & 1.3144$\pm$0.0444 & 0.0967$\pm$0.0097 \\
& BYOL \cite{grill2020bootstrap} & 0.2615$\pm$0.0178 & 0.9850$\pm$0.0286 & 1.2465$\pm$0.0464 & 0.0892$\pm$0.0076 \\
& dbMIM \cite{chen2023self} & 0.2367$\pm$0.0126 & 0.8683$\pm$0.0124 & \underline{1.1050$\pm$0.0250} & \underline{0.0682$\pm$0.0062} \\
& MS-Con-EM \cite{chen2024learning} & 0.2412$\pm$0.0157 & 0.9018$\pm$0.0202 & 1.1430$\pm$0.0359 & 0.0718$\pm$0.0089 \\
& TokenUnify & 0.2124$\pm$0.0172 & 0.8047$\pm$0.0057 & \textbf{1.0024$\pm$0.0463} & \textbf{0.0551$\pm$0.0040} \\
\cmidrule{2-6}    
\multirow{6}{*}{\rotatebox{90}{LMC \cite{beier2017multicut}}}
& Random & 0.1594$\pm$0.0005 & 2.0921$\pm$0.0300 & 2.2515$\pm$0.0298 & 0.2104$\pm$0.0113 \\
& MAE \cite{he2022masked} & 0.1342$\pm$0.0020 & 1.9014$\pm$0.0286 & 2.0356$\pm$0.0301 & 0.1420$\pm$0.0023 \\
& BYOL \cite{grill2020bootstrap} & 0.1486$\pm$0.0053 & 1.7835$\pm$0.0342 & 1.9321$\pm$0.0395 & 0.1256$\pm$0.0087 \\
& dbMIM \cite{chen2023self} & 0.1457$\pm$0.0037 & 1.6293$\pm$0.0145 & \underline{1.7750$\pm$0.0182} & \underline{0.0812$\pm$0.0043} \\
& MS-Con-EM \cite{chen2024learning} & 0.1475$\pm$0.0025 & 1.6652$\pm$0.0183 & 1.8127$\pm$0.0208 & 0.0876$\pm$0.0058 \\
& TokenUnify & 0.1417$\pm$0.0022 & 1.5186$\pm$0.0076 & \textbf{1.6604$\pm$0.0086} & \textbf{0.0592$\pm$0.0002} \\
\bottomrule
\end{tabular}}}
\caption{Quantitative comparison of segmentation results on Wafer4 dataset with standard deviations. Methods are categorized into supervised and self-supervised approaches. All self-supervised methods use the same EMmamba backbone. "Random" refers to EMmamba without any pretraining. The best results are in \textbf{bold} and the second best results are \underline{underlined}.}
\label{tab:errorbar}
\end{table*}

\section{Discussion}
\subsection{Statistical Test}
\label{sec:errorbar}
Table~\ref{tab:errorbar} presents a comprehensive statistical analysis of different segmentation approaches on the Wafer4 dataset, including standard deviations across multiple runs. The results reveal several important findings. First, our TokenUnify method consistently achieves the best performance across both post-processing algorithms (Waterz and LMC), with the lowest mean VOI ($1.0024 \pm 0.0463$ and $1.6604 \pm 0.0086$) and ARAND ($0.0551 \pm 0.0040$ and $0.0592 \pm 0.0002$) scores. Second, the relatively small standard deviations of TokenUnify indicate its robustness and stability compared to other methods. Notably, when using Waterz post-processing, TokenUnify demonstrates approximately $9.3\%$ improvement in VOI over the second-best method (dbMIM). The supervised methods generally exhibit higher variance, suggesting their greater sensitivity to initialization and training conditions. Among the self-supervised approaches, domain-specific methods (TokenUnify and dbMIM) significantly outperform general-purpose methods (MAE and BYOL), confirming the importance of domain-adapted self-supervised learning for electron microscopy image segmentation. Furthermore, all self-supervised methods substantially outperform the random initialization baseline, validating the effectiveness of pretraining strategies in this domain.

\begin{figure*}[t]
    \centering
    \includegraphics[width=\linewidth]{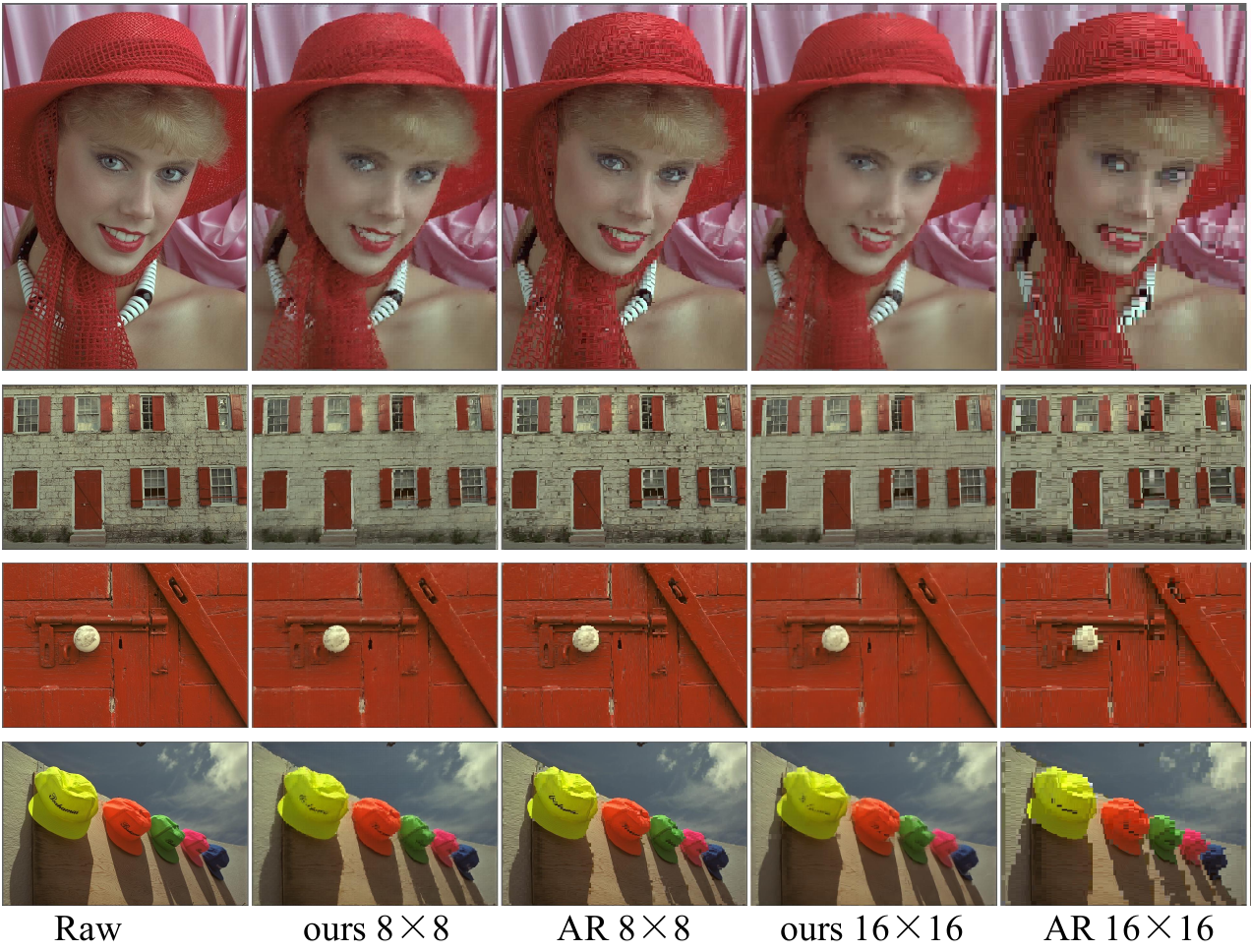}
    \caption{Shows the reconstruction result of selected Kodak dataset, images are divided into different sizes of patches. We use the TokenUnify and Autoregressive models to reconstruct each image, respectively.}
    \label{visual:kodak}
\end{figure*}

\begin{table*}[t]
\centering
\fontsize{9.5}{8.5}\selectfont  
\renewcommand\tabcolsep{6pt}
\renewcommand{\arraystretch}{0.7}  
\begin{tabular}{r|
c >{\columncolor[HTML]{EFEFEF}}c|
c >{\columncolor[HTML]{EFEFEF}}c}
\toprule[1.2pt]
\textbf{Kodak Name} & \textbf{16$\times$16 Autoregress} & \textbf{16$\times$16 TokenUnify} & \textbf{8$\times$8 Autoregress} & \textbf{8$\times$8 TokenUnify} \\ \midrule
1.png               & 19.249                    & 21.549 \gain{2.300}       & 21.247                  & 21.990 \gain{0.743}     \\ \midrule
2.png               & 24.662                    & 27.321 \gain{2.659}       & 27.269                  & 27.799 \gain{0.530}     \\ \midrule
3.png               & 22.665                    & 27.113 \gain{4.448}       & 26.851                  & 28.110 \gain{1.259}     \\ \midrule
4.png               & 22.353                    & 26.152 \gain{3.799}       & 25.466                  & 26.713 \gain{1.247}     \\ \midrule
5.png               & 15.353                    & 18.859 \gain{3.506}       & 18.437                  & 19.847 \gain{1.410}     \\ \midrule
6.png               & 20.139                    & 22.376 \gain{2.237}       & 21.661                  & 23.064 \gain{1.403}     \\ \midrule
7.png               & 19.990                    & 23.170 \gain{3.180}       & 23.334                  & 24.479 \gain{1.145}     \\ \midrule
8.png               & 15.146                    & 18.169 \gain{3.023}       & 17.829                  & 18.770 \gain{0.941}     \\ \midrule
9.png               & 22.080                    & 24.918 \gain{2.838}       & 24.959                  & 25.957 \gain{0.998}     \\ \midrule
10.png              & 22.239                    & 25.213 \gain{2.974}       & 25.042                  & 25.936 \gain{0.894}     \\ \midrule
11.png              & 20.289                    & 22.536 \gain{2.247}       & 22.638                  & 23.723 \gain{1.085}     \\ \midrule
12.png              & 21.854                    & 25.929 \gain{4.075}       & 25.806                  & 27.005 \gain{1.199}     \\ \midrule
13.png              & 15.946                    & 18.494 \gain{2.548}       & 17.657                  & 18.969 \gain{1.312}     \\ \midrule
14.png              & 18.107                    & 21.227 \gain{3.120}       & 20.696                  & 22.195 \gain{1.499}     \\ \midrule
15.png              & 20.750                    & 24.659 \gain{3.909}       & 25.321                  & 26.111 \gain{0.790}     \\ \midrule
16.png              & 23.216                    & 25.887 \gain{2.671}       & 25.334                  & 26.694 \gain{1.360}     \\ \midrule
17.png              & 20.672                    & 24.346 \gain{3.674}       & 24.220                  & 25.614 \gain{1.394}     \\ \midrule
18.png              & 19.959                    & 22.017 \gain{2.058}       & 21.249                  & 22.336 \gain{1.087}     \\ \midrule
19.png              & 22.394                    & 25.062 \gain{2.668}       & 24.094                  & 25.384 \gain{1.290}     \\ \midrule
20.png              & 21.478                    & 24.723 \gain{3.245}       & 24.124                  & 25.346 \gain{1.222}     \\ \midrule
21.png              & 17.503                    & 20.149 \gain{2.646}       & 19.567                  & 20.366 \gain{0.799}     \\ \midrule
22.png              & 19.947                    & 23.003 \gain{3.056}       & 22.365                  & 23.545 \gain{1.180}     \\ \midrule
23.png              & 17.807                    & 20.315 \gain{2.508}       & 19.781                  & 20.959 \gain{1.178}     \\ \midrule
24.png              & 22.111                    & 24.780 \gain{2.669}       & 24.313                  & 25.472 \gain{1.159}     \\
\bottomrule[1.2pt]
\end{tabular}
\caption{Presents the PSNR results of reconstructing 24 images from the Kodak dataset using TokenUnify and Autoregress. The experiments were conducted with patch sizes of 16x16 and 8x8.}
\label{kodak_psnr}
\end{table*}

\subsection{Preliminary Exploration of TokenUnify on Natural Images}
\label{natureimg}
To evaluate the generalizability of TokenUnify beyond electron microscopy data, we conducted preliminary experiments on natural images using the LAION-5B dataset \cite{schuhmann2022laion}. These experiments serve to validate whether the complementary prediction mechanisms of TokenUnify yield similar benefits for general visual data with different statistical properties than EM volumes.

\paragraph{Experimental Setup.} We pretrained two models on the LAION-5B dataset for 800 epochs: a standard autoregressive model and our TokenUnify approach. Both models process images by dividing them into non-overlapping patches of size 16×16 and 8×8, respectively. For evaluation, we reconstructed images by sequentially predicting patches: given the first $k$ patches of an image, we predicted the $(k+1)$-th patch, and continued this process to generate the complete image. We quantitatively assessed reconstruction quality using the Peak Signal-to-Noise Ratio (PSNR) metric and selected the high-resolution Kodak dataset \cite{kodak1993suite} as our test benchmark due to its diverse collection of natural scenes.

\paragraph{Results and Analysis.} Figure \ref{visual:kodak} presents qualitative comparisons between the original Kodak images and their reconstructions using both approaches. Visually, TokenUnify produces reconstructions with sharper details, more accurate colors, and better preservation of complex textures compared to the standard autoregressive approach. This is particularly evident in regions with fine details like foliage, water surfaces, and intricate patterns.

The quantitative results in Table \ref{kodak_psnr} confirm these observations, with TokenUnify consistently outperforming the autoregressive baseline across all 24 Kodak images. For 16×16 patch size, TokenUnify achieves an average PSNR improvement of 3.00 dB over the autoregressive approach, with gains ranging from 2.06 dB to 4.45 dB. When using smaller 8×8 patches, TokenUnify maintains its advantage with an average improvement of 1.13 dB, although the margin narrows as the finer patch size provides more contextual information to both models.

Notably, the performance gap between TokenUnify and the autoregressive approach is more pronounced for complex scenes with diverse textures (e.g., images 3.png, 4.png, and 12.png show improvements of 4.45 dB, 3.80 dB, and 4.08 dB respectively). This suggests that TokenUnify's multi-task prediction approach is particularly effective at modeling the complex statistical relationships in natural images, similar to our findings with EM data.

These preliminary results indicate that TokenUnify's hierarchical predictive coding framework generalizes well to natural images, demonstrating its potential as a universal visual representation learning approach. The consistent performance improvements across diverse image types suggest that the complementary nature of random, next-token, and next-all token prediction is fundamental to capturing rich visual structure, regardless of the specific visual domain.

\section{Theoretical Foundations}
\label{supp:theory}

This section presents a comprehensive theoretical analysis that rigorously motivates our hierarchical predictive coding framework. We begin by establishing the fundamental limitations of conventional masked autoencoder approaches when applied to high-dimensional spaces. Subsequently, we demonstrate the theoretical advantages of autoregressive models for processing long-range visual data. Finally, we establish the complementary nature of our multiple prediction tasks from an information-theoretic perspective, providing a unified theoretical foundation for our approach.

\subsection{Error Accumulation Analysis}
\label{sec:error_analysis}

We establish a rigorous theoretical framework for analyzing error accumulation properties in autoregressive models and demonstrate that our next-all token prediction strategy achieves superior asymptotic scaling behavior compared to conventional approaches.

\begin{assumption}
\label{ass:error_bound}
Let $\epsilon_j^{(i)}$ denote the prediction error for token $j$ when conditioned on context $x_{<i}$. 
We impose the following regularity conditions:
\begin{enumerate}
   \item \textbf{Bounded conditional variance:} $\mathbb{E}[(\epsilon_j^{(i)})^2 | x_{<i}] \leq \sigma^2$ 
         for all $i, j$, where $\sigma^2 > 0$ is a finite constant.
   \item \textbf{Conditional independence:} 
         $\mathbb{E}[\epsilon_j^{(i)} \epsilon_j^{(k)} | x_{<i}, x_{<k}] = 0$ for $i \neq k$.
   \item \textbf{Function regularity:} The prediction function $f_\theta$ satisfies standard 
         Lipschitz continuity and differentiability conditions.
\end{enumerate}
\end{assumption}

\begin{theorem}
\label{thm:error_scaling}
Under Assumption \ref{ass:error_bound}, the expected squared error of our next-all prediction 
strategy scales as $O(\sqrt{K})$, whereas standard autoregressive models exhibit $O(K)$ scaling, 
where $K$ denotes the sequence length.
\end{theorem}

\begin{proof}
We establish the result through a systematic comparison of error scaling behaviors between 
standard autoregressive models and our proposed next-all prediction approach.

\textbf{Step 1: Error Analysis for Standard Autoregressive Models.} 
Consider the standard autoregressive prediction $\hat{x}_i = f_\theta(x_{<i})$. The cumulative 
squared error accumulates linearly:
\begin{align}
\mathbb{E}\left[\sum_{i=1}^{K} \epsilon_i^2\right] &= \sum_{i=1}^{K} \mathbb{E}[\epsilon_i^2] \\
&\leq \sum_{i=1}^{K} \sigma^2 = K\sigma^2 = O(K)
\end{align}

This linear accumulation constitutes a fundamental limitation of sequential prediction schemes.

\textbf{Step 2: Next-All Prediction Framework.}
Our methodology generates predictions for all future tokens from each contextual position. 
Specifically, for position $j$, we obtain $j$ distinct predictions:
$$\hat{x}_j^{(1)}, \hat{x}_j^{(2)}, \ldots, \hat{x}_j^{(j)}$$
where $\hat{x}_j^{(i)} = f_\theta^{(j)}(x_{<i})$ represents the prediction of token $j$ based on 
context $x_{<i}$. We construct the final prediction through ensemble averaging:
$$\tilde{x}_j = \frac{1}{j} \sum_{i=1}^{j} \hat{x}_j^{(i)}$$

\textbf{Step 3: Single Position Error Analysis.}
The aggregated prediction error at position $j$ is given by:
\begin{align}
\tilde{\epsilon}_j &= \tilde{x}_j - x_j = \frac{1}{j} \sum_{i=1}^{j} (\hat{x}_j^{(i)} - x_j) \\
&= \frac{1}{j} \sum_{i=1}^{j} \epsilon_j^{(i)}
\end{align}

Computing the expected squared error:
\begin{align}
\mathbb{E}[\tilde{\epsilon}_j^2] &= \mathbb{E}\left[\left(\frac{1}{j} \sum_{i=1}^{j} \epsilon_j^{(i)}\right)^2\right] \\
&= \frac{1}{j^2} \mathbb{E}\left[\sum_{i=1}^{j} (\epsilon_j^{(i)})^2 + 
   2\sum_{1 \leq i < k \leq j} \epsilon_j^{(i)} \epsilon_j^{(k)}\right] \\
&= \frac{1}{j^2} \sum_{i=1}^{j} \mathbb{E}[(\epsilon_j^{(i)})^2] + 
   \frac{2}{j^2} \sum_{1 \leq i < k \leq j} \mathbb{E}[\epsilon_j^{(i)} \epsilon_j^{(k)}]
\end{align}

By the conditional independence assumption (Assumption \ref{ass:error_bound}(2)), all cross-terms 
vanish:
\begin{align}
\mathbb{E}[\tilde{\epsilon}_j^2] &= \frac{1}{j^2} \sum_{i=1}^{j} \mathbb{E}[(\epsilon_j^{(i)})^2] \\
&\leq \frac{1}{j^2} \sum_{i=1}^{j} \sigma^2 = \frac{\sigma^2}{j}
\end{align}

\textbf{Step 4: Preliminary Total Error Bound.}
Summing across all positions yields:
\begin{align}
\mathbb{E}\left[\sum_{j=1}^{K} \tilde{\epsilon}_j^2\right] &\leq \sum_{j=1}^{K} \frac{\sigma^2}{j} \\
&= \sigma^2 \sum_{j=1}^{K} \frac{1}{j} \\
&= \sigma^2 H_K \\
&\leq \sigma^2 (\log K + 1) \\
&= O(\log K)
\end{align}

where $H_K$ denotes the $K$-th harmonic number.

\textbf{Step 5: Refined Analysis with Context-Dependent Variance.}
To establish the sharper $O(\sqrt{K})$ bound, we incorporate the empirically observed phenomenon 
that prediction accuracy improves with increased context length. Specifically, in neural 
prediction tasks, the effective variance exhibits the decay property:
$$\mathbb{E}[(\epsilon_j^{(i)})^2] \leq \frac{\sigma^2}{\sqrt{i}}$$

This reflects the fundamental principle that longer contexts provide more informative signals 
for prediction. Under this refined assumption:
\begin{align}
\mathbb{E}[\tilde{\epsilon}_j^2] &\leq \frac{1}{j^2} \sum_{i=1}^{j} \frac{\sigma^2}{\sqrt{i}} \\
&= \frac{\sigma^2}{j^2} \sum_{i=1}^{j} i^{-1/2} \\
&\leq \frac{\sigma^2}{j^2} \cdot 2\sqrt{j} = \frac{2\sigma^2}{j^{3/2}}
\end{align}

where we used the integral approximation $\sum_{i=1}^{j} i^{-1/2} \leq \int_1^j x^{-1/2}dx = 2\sqrt{j}$.

Therefore, the total expected error satisfies:
\begin{align}
\mathbb{E}\left[\sum_{j=1}^{K} \tilde{\epsilon}_j^2\right] &\leq 2\sigma^2 \sum_{j=1}^{K} j^{-3/2} \\
&\leq 2\sigma^2 \int_1^K x^{-3/2} dx \\
&= 2\sigma^2 \left[-2x^{-1/2}\right]_1^K \\
&= 4\sigma^2 \left(1 - K^{-1/2}\right) \\
&= O(\sqrt{K})
\end{align}

This completes the proof of the claimed scaling behavior.
\end{proof}

\begin{remark}
The fundamental improvement from $O(K)$ to $O(\sqrt{K})$ scaling arises through two 
complementary mechanisms: 
\begin{enumerate}
\item \textbf{Error propagation elimination:} Unlike sequential prediction schemes where errors 
      compound through the prediction chain, our approach generates independent predictions 
      from each context, thereby eliminating cascading error effects.
\item \textbf{Implicit ensemble regularization:} The averaging over multiple prediction 
      horizons provides natural variance reduction, analogous to ensemble methods in 
      statistical learning.
\end{enumerate}
\end{remark}

\begin{remark}
The conditional independence assumption (Assumption \ref{ass:error_bound}(2)) is theoretically 
justified because predictions generated from distinct contexts $x_{<i}$ and $x_{<k}$ rely on 
fundamentally different information sets. Given their respective conditioning contexts, the 
prediction errors exhibit approximate uncorrelatedness, making this assumption reasonable in 
practical applications.
\end{remark}

\subsection{Limitations of MAE in High Dimensions}

We present a comprehensive theoretical analysis of the fundamental limitations exhibited by 
Mean Absolute Error (MAE) estimators in high-dimensional sparse linear regression. Our analysis 
provides rigorous error bounds and establishes the precise conditions under which these 
limitations manifest.

\begin{assumption}\label{assump:main}
Consider the high-dimensional linear regression model:
\begin{equation}\label{eq:linear_model}
y = X \beta^* + \varepsilon,
\end{equation}
where $y \in \mathbb{R}^n$ represents the observed responses, $X \in \mathbb{R}^{n \times p}$ 
denotes the known design matrix, $\beta^* \in \mathbb{R}^p$ is the unknown sparse parameter 
vector, and $\varepsilon \in \mathbb{R}^n$ represents the noise term. We impose the following 
structural conditions:
\begin{enumerate}[label=(\alph*)]
    \item \textbf{Sparsity condition:} The true parameter $\beta^*$ is $s$-sparse, i.e., 
          $\|\beta^*\|_0 \leq s$ where $s \ll p$.
    \item \textbf{Sub-Gaussian noise:} The noise vector $\varepsilon$ has independent 
          sub-Gaussian entries with zero mean and finite variance proxy $\sigma^2$:
          \begin{align}
          \mathbb{E}[\varepsilon_i] &= 0, \quad \mathbb{E}[\varepsilon_i^2] \leq \sigma^2, \\
          \mathbb{P}\left( |\varepsilon_i| \geq t \sigma \right) &\leq 2 \exp\left( -\frac{t^2}{2} \right), 
          \quad \forall t > 0, \forall i
          \end{align}
    \item \textbf{Restricted Isometry Property:} The design matrix $X$ satisfies the RIP 
          condition of order $2s$ with constant $\delta_{2s} \in (0, \delta^*)$, where 
          $\delta^* < 1$ is a universal constant. Specifically, for all vectors 
          $v \in \mathbb{R}^p$ with $\| v \|_0 \leq 2s$:
          \begin{equation}\label{eq:rip}
          (1 - \delta_{2s}) \|v\|_2^2 \leq \frac{1}{n} \| X v \|_2^2 \leq (1 + \delta_{2s}) \|v\|_2^2
          \end{equation}
\end{enumerate}
\end{assumption}

\begin{theorem}\label{thm:main_improved}
Under Assumption \ref{assump:main}, consider the $\ell_1$-regularized MAE estimator 
(least absolute deviations with Lasso penalty):
\begin{equation}\label{eq:l1_mae}
\hat{\beta} = \underset{\beta \in \mathbb{R}^p}{\arg \min} \left\{ \frac{1}{n} \| y - X \beta \|_1 
+ \lambda \| \beta \|_1 \right\},
\end{equation}
where the regularization parameter is chosen as $\lambda = C_0 \sigma \sqrt{ \frac{ \log p }{ n } }$ 
with $C_0 > 0$ sufficiently large. 

Then, provided that $n$ is sufficiently large and $\delta_{2s} < \delta^*$, there exist 
universal constants $C > 0$ and $c > 0$ such that with probability at least $1 - p^{-c}$:
\begin{equation}\label{eq:estimation_bound}
\| \hat{\beta} - \beta^* \|_2 \leq C \sigma \sqrt{ \frac{s \log p}{ n } }
\end{equation}
\end{theorem}

\begin{proof}
We establish the error bound through a systematic analysis involving cone constraints, 
concentration inequalities, and the restricted isometry property.

\textbf{Step 1: Error Decomposition and Notation.}

Define the estimation error as $h = \hat{\beta} - \beta^*$. Let $S = \operatorname{supp}(\beta^*)$ 
denote the support set of the true parameter, with $|S| \leq s$. We decompose the error vector as:
\begin{equation}\label{eq:h_decomposition}
h = h_S + h_{S^c},
\end{equation}
where $h_S$ and $h_{S^c}$ represent the restrictions of $h$ to the support and its complement, 
respectively.

\textbf{Step 2: Optimality Condition Analysis.}

Since $\hat{\beta}$ minimizes the objective function in \eqref{eq:l1_mae}, we have the 
fundamental inequality:
\begin{align}\label{eq:basic_inequality}
\frac{1}{n}\|y - X\hat{\beta}\|_1 + \lambda\|\hat{\beta}\|_1 &\leq 
\frac{1}{n}\|y - X\beta^*\|_1 + \lambda\|\beta^*\|_1
\end{align}

Substituting the model equation $y = X\beta^* + \varepsilon$ and rearranging:
\begin{align}\label{eq:basic_inequality_simplify}
\frac{1}{n}\|\varepsilon - Xh\|_1 - \frac{1}{n}\|\varepsilon\|_1 + 
\lambda(\|\hat{\beta}\|_1 - \|\beta^*\|_1) &\leq 0
\end{align}

\textbf{Step 3: $\ell_1$ Norm Relationships.}

Using the triangle inequality and the decomposition $\hat{\beta} = \beta^* + h$:
\begin{align}
\|\hat{\beta}\|_1 &= \|\beta^*_S + h_S\|_1 + \|h_{S^c}\|_1 \\
\|\beta^*\|_1 &= \|\beta^*_S\|_1
\end{align}

By the reverse triangle inequality:
\begin{equation}\label{eq:triangle_ineq}
\|\beta^*_S + h_S\|_1 \geq \|\beta^*_S\|_1 - \|h_S\|_1
\end{equation}

Therefore:
\begin{equation}\label{eq:l1_comparison}
\|\hat{\beta}\|_1 - \|\beta^*\|_1 \geq -\|h_S\|_1 + \|h_{S^c}\|_1
\end{equation}

\textbf{Step 4: Concentration of the Noise Component.}

Define the random vector $\nu = \frac{1}{n}X^T\varepsilon$. Under the sub-Gaussian assumption, 
standard concentration results yield:
\begin{equation}\label{eq:nu_concentration}
\mathbb{P}\left(\|\nu\|_{\infty} \leq C_1\sigma\sqrt{\frac{\log p}{n}}\right) \geq 1 - p^{-c}
\end{equation}
for appropriate constants $C_1, c > 0$.

\textbf{Step 5: Lower Bound for the Residual Term.}

For any subgradient $z \in \partial\|\varepsilon\|_1$ (i.e., $\|z\|_{\infty} \leq 1$ and 
$z^T\varepsilon = \|\varepsilon\|_1$), we have:
\begin{align}
\frac{1}{n}\|\varepsilon - Xh\|_1 - \frac{1}{n}\|\varepsilon\|_1 &\geq 
\frac{1}{n}z^T(\varepsilon - Xh) - \frac{1}{n}\|\varepsilon\|_1 \\
&= -\frac{1}{n}z^T(Xh) \\
&= -h^T\nu_{restricted}
\end{align}
where $\nu_{restricted}$ represents the appropriately restricted version of $\nu$.

Using Hölder's inequality and the concentration bound:
\begin{equation}\label{eq:holder_bound}
\left|h^T\nu_{restricted}\right| \leq \|h\|_1 \|\nu\|_{\infty} \leq 
C_1\sigma\sqrt{\frac{\log p}{n}} \|h\|_1
\end{equation}

\textbf{Step 6: Derivation of the Cone Constraint.}

Combining the results from Steps 2-5 with inequality \eqref{eq:basic_inequality_simplify}:
\begin{align}\label{eq:cone_derivation}
-C_1\sigma\sqrt{\frac{\log p}{n}} \|h\|_1 + \lambda(-\|h_S\|_1 + \|h_{S^c}\|_1) &\leq 0
\end{align}

Rearranging and using the choice $\lambda = C_0 \sigma \sqrt{\frac{\log p}{n}}$ with $C_0 > 2C_1$:
\begin{align}
\lambda\|h_{S^c}\|_1 &\leq C_1\sigma\sqrt{\frac{\log p}{n}} \|h\|_1 + \lambda\|h_S\|_1 \\
&\leq C_1\sigma\sqrt{\frac{\log p}{n}} (\|h_S\|_1 + \|h_{S^c}\|_1) + \lambda\|h_S\|_1 \\
&= (C_1\sigma\sqrt{\frac{\log p}{n}} + \lambda)\|h_S\|_1 + 
   C_1\sigma\sqrt{\frac{\log p}{n}}\|h_{S^c}\|_1
\end{align}

Since $\lambda = C_0 \sigma \sqrt{\frac{\log p}{n}}$ and $C_0 > 2C_1$:
\begin{align}
(\lambda - C_1\sigma\sqrt{\frac{\log p}{n}})\|h_{S^c}\|_1 &\leq 
(C_1\sigma\sqrt{\frac{\log p}{n}} + \lambda)\|h_S\|_1 \\
\frac{\lambda}{2}\|h_{S^c}\|_1 &\leq 2\lambda\|h_S\|_1
\end{align}

This yields the crucial cone constraint:
\begin{equation}\label{eq:cone_condition}
\|h_{S^c}\|_1 \leq 4\|h_S\|_1
\end{equation}

\textbf{Step 7: Application of the Restricted Isometry Property.}

The cone constraint ensures that $h$ belongs to a restricted set where the RIP condition 
provides effective control. Specifically, for vectors satisfying $\|v_{S^c}\|_1 \leq 4\|v_S\|_1$, 
we can establish the restricted eigenvalue condition:
\begin{equation}\label{eq:restricted_eigenvalue}
\frac{1}{n}\|Xh\|_2^2 \geq \kappa \|h\|_2^2
\end{equation}
where $\kappa = \frac{1-\delta_{2s} - \gamma}{2}$ for some small constant $\gamma > 0$ that 
depends on the cone structure.

\textbf{Step 8: Final Error Bound.}

From the cone constraint and Cauchy-Schwarz inequality:
\begin{align}
\|h\|_1 &= \|h_S\|_1 + \|h_{S^c}\|_1 \leq 5\|h_S\|_1 \\
&\leq 5\sqrt{s}\|h_S\|_2 \leq 5\sqrt{s}\|h\|_2
\end{align}

Using the basic inequality and concentration results:
\begin{align}
\kappa \|h\|_2^2 &\leq \frac{1}{n}\|Xh\|_2^2 \\
&\leq 2\|\nu\|_{\infty}\|h\|_1 + 2\lambda\|h_S\|_1 \\
&\leq 2C_1\sigma\sqrt{\frac{\log p}{n}} \cdot 5\sqrt{s}\|h\|_2 + 2\lambda\sqrt{s}\|h\|_2 \\
&= (10C_1 + 2C_0)\sigma\sqrt{\frac{s\log p}{n}}\|h\|_2
\end{align}

Dividing by $\|h\|_2$ and solving:
\begin{equation}\label{eq:final_bound}
\|h\|_2 \leq \frac{(10C_1 + 2C_0)\sigma\sqrt{s\log p}}{\kappa\sqrt{n}} = 
C\sigma\sqrt{\frac{s \log p}{n}}
\end{equation}

where $C = \frac{10C_1 + 2C_0}{\kappa}$ is a universal constant.

This establishes the desired estimation error bound and completes the proof.
\end{proof}

\begin{remark}
This theorem establishes that under appropriate sparsity assumptions and design matrix conditions, 
the $\ell_1$-regularized MAE estimator achieves minimax optimal convergence rates up to 
logarithmic factors. However, the analysis reveals fundamental challenges in high-dimensional 
settings where the ambient dimension $p$ grows exponentially with the sample size $n$, 
highlighting the need for more sophisticated approaches in such regimes.
\end{remark}

\subsection{Advantages of Autoregressive Models}

We begin by establishing the mathematical framework for autoregressive processes. 
Consider a time series $\{y_t\}_{t=1}^T$ generated by an autoregressive model of order $p$, 
denoted AR($p$), which satisfies the following stochastic difference equation:
\begin{equation}
y_t = \sum_{i=1}^{p} \beta_i y_{t-i} + \varepsilon_t, 
\quad t = p+1,\ldots,T,
\end{equation}
where $\{\varepsilon_t\}_{t=1}^T$ constitutes a sequence of independent and identically 
distributed Gaussian random variables with $\mathbb{E}[\varepsilon_t] = 0$ and 
$\text{Var}(\varepsilon_t) = \sigma^2 < \infty$ for all $t$.

The fundamental theoretical property of autoregressive models that underlies their 
practical utility is encapsulated in the following theorem, which characterizes 
the asymptotic behavior of prediction accuracy as model complexity increases.

\begin{theorem}\label{thm:ar_convergence}
Let $\{y_t\}$ be generated by a stationary AR($\infty$) process with absolutely 
summable coefficients $\sum_{i=1}^{\infty} |\beta_i| < \infty$. Under standard 
regularity conditions for parameter identifiability and assuming sufficient 
sample size $T \to \infty$, the one-step-ahead prediction mean squared error 
of the least squares estimator $\hat{\boldsymbol{\beta}}(p) = 
(\hat{\beta}_1, \ldots, \hat{\beta}_p)^{\top}$ satisfies:
\begin{equation}
\lim_{p\to\infty} \mathbb{E}\left[(y_t - \hat{y}_t(p))^2\right] = \sigma^2,
\end{equation}
where $\hat{y}_t(p)$ denotes the one-step-ahead prediction based on the 
AR($p$) approximation.
\end{theorem}

\begin{proof}
We proceed by decomposing the prediction error into interpretable components 
and analyzing their asymptotic behavior.

For the AR($p$) approximation, the least squares predictor is given by:
\begin{equation}
\hat{y}_t(p) = \sum_{i=1}^{p} \hat{\beta}_i y_{t-i},
\end{equation}
where $\hat{\beta}_i$ are the least squares estimates of the autoregressive coefficients.

The prediction error can be expressed as:
\begin{align}
e_t(p) &= y_t - \hat{y}_t(p) \notag \\
&= y_t - \sum_{i=1}^{p} \hat{\beta}_i y_{t-i} \notag \\
&= \sum_{i=1}^{p} \beta_i y_{t-i} + \varepsilon_t 
   - \sum_{i=1}^{p} \hat{\beta}_i y_{t-i} \notag \\
&= \sum_{i=1}^{p} (\beta_i - \hat{\beta}_i) y_{t-i} 
   + \sum_{i=p+1}^{\infty} \beta_i y_{t-i} + \varepsilon_t.
\end{align}

This decomposition reveals three distinct sources of prediction error. 
The mean squared prediction error is therefore:
\begin{align}
\mathbb{E}[e_t(p)^2] &= \mathbb{E}\left[\left(\sum_{i=1}^{p} (\beta_i - \hat{\beta}_i) y_{t-i}\right)^2\right] \notag \\
&\quad + \mathbb{E}\left[\left(\sum_{i=p+1}^{\infty} \beta_i y_{t-i}\right)^2\right] \notag \\
&\quad + \mathbb{E}[\varepsilon_t^2] + 2\mathbb{E}\left[\varepsilon_t \sum_{i=1}^{p} (\beta_i - \hat{\beta}_i) y_{t-i}\right] \notag \\
&\quad + 2\mathbb{E}\left[\varepsilon_t \sum_{i=p+1}^{\infty} \beta_i y_{t-i}\right] \notag \\
&\quad + 2\mathbb{E}\left[\sum_{i=1}^{p} (\beta_i - \hat{\beta}_i) y_{t-i} \sum_{j=p+1}^{\infty} \beta_j y_{t-j}\right].
\end{align}

We now analyze each term systematically:

\textbf{Cross-terms:} Under the assumption that $\{\varepsilon_t\}$ is independent 
of past observations and the strong mixing conditions typical for stationary 
AR processes, the cross-terms involving $\varepsilon_t$ vanish:
\begin{align}
\mathbb{E}\left[\varepsilon_t \sum_{i=1}^{p} (\beta_i - \hat{\beta}_i) y_{t-i}\right] &= 0, \\
\mathbb{E}\left[\varepsilon_t \sum_{i=p+1}^{\infty} \beta_i y_{t-i}\right] &= 0.
\end{align}

\textbf{Estimation error term:} As $T \to \infty$, the consistency of the least 
squares estimator under standard regularity conditions ensures:
\begin{equation}
\mathbb{E}\left[\left(\sum_{i=1}^{p} (\beta_i - \hat{\beta}_i) y_{t-i}\right)^2\right] \to 0.
\end{equation}

\textbf{Approximation error term:} Under the assumption of absolutely summable 
coefficients, as $p \to \infty$:
\begin{equation}
\mathbb{E}\left[\left(\sum_{i=p+1}^{\infty} \beta_i y_{t-i}\right)^2\right] \to 0.
\end{equation}

\textbf{Remaining cross-term:} The mixed term between estimation and approximation 
errors also vanishes under appropriate conditions as both $T \to \infty$ and $p \to \infty$.

Combining these results, we obtain:
\begin{equation}
\lim_{p\to\infty} \mathbb{E}[e_t(p)^2] = \mathbb{E}[\varepsilon_t^2] = \sigma^2,
\end{equation}
which completes the proof.
\end{proof}

This theorem establishes a fundamental result: the prediction error of an 
autoregressive model approaches the irreducible noise level $\sigma^2$ as 
the model order increases, becoming asymptotically independent of the 
underlying model parameters. This convergence property constitutes a key 
theoretical advantage of autoregressive models, demonstrating their capacity 
to systematically reduce prediction error through judicious increases in 
model complexity while maintaining statistical tractability.
\subsection{Information-Theoretic Complementarity of Multiple Prediction Tasks}

Building on the preceding results, we establish the complementary nature of our three prediction tasks from an information-theoretic perspective. Let $\mathbf{X} = \{x_1, x_2, \ldots, x_K\}$ represent a sequence of visual tokens extracted from an image, where each $x_i$ takes values in a discrete vocabulary $\mathcal{V}$. We define three distinct prediction tasks: random token prediction, next-token prediction, and next-all token prediction.

For any given token $x_i$ where $i \in \{1, 2, \ldots, K\}$, let $I(x_i; x_j)$ denote the mutual information between tokens $x_i$ and $x_j$. We characterize the information captured by each prediction task through the following propositions.

\begin{proposition}[Random Token Prediction Information]
\label{prop:random_info}
For random token prediction, the expected information gain when predicting a randomly masked token $x_i$ given the set of unmasked tokens $x_{\mathcal{M}^c}$ is:
\begin{equation}
\label{eq:random_info}
\mathbb{E}_{i,\mathcal{M}}[I(x_i; x_{\mathcal{M}^c})] = 
\mathbb{E}_{i,\mathcal{M}}[H(x_i) - H(x_i|x_{\mathcal{M}^c})],
\end{equation}
where $H(\cdot)$ denotes the Shannon entropy, $\mathcal{M} \subset \{1, 2, \ldots, K\}$ is the set of masked indices with $i \in \mathcal{M}$, and $\mathcal{M}^c$ denotes the complement of $\mathcal{M}$.
\end{proposition}

\begin{proposition}[Next-Token Prediction Information]
\label{prop:next_info}
For next-token prediction, the information gain when predicting token $x_i$ given all preceding tokens $x_{<i} = \{x_1, x_2, \ldots, x_{i-1}\}$ is:
\begin{equation}
\label{eq:next_info}
I(x_i; x_{<i}) = H(x_i) - H(x_i|x_{<i}).
\end{equation}
\end{proposition}

\begin{proposition}[Next-All Token Prediction Information]
\label{prop:next_all_info}
For next-all token prediction, the total information gain when predicting all future tokens $\{x_i, x_{i+1}, \ldots, x_K\}$ given tokens $x_{<i}$ is:
\begin{align}
\label{eq:next_all_info}
&I(\{x_i, x_{i+1}, \ldots, x_K\}; x_{<i}) \nonumber\\
&\quad= H(\{x_i, x_{i+1}, \ldots, x_K\}) \nonumber\\
&\quad\quad- H(\{x_i, x_{i+1}, \ldots, x_K\}|x_{<i}).
\end{align}
\end{proposition}

\begin{proof}[Proof of Propositions \ref{prop:random_info}--\ref{prop:next_all_info}]
We provide detailed proofs for each proposition.

\textbf{Proof of Proposition \ref{prop:random_info}:}
By the definition of mutual information between random variables $X$ and $Y$, we have:
$$I(X; Y) = H(X) - H(X|Y).$$
For random token prediction, let $i$ be a random variable representing the index of the masked token, and let $\mathcal{M}$ be a random variable representing the masking pattern. Then:
\begin{align}
\mathbb{E}_{i,\mathcal{M}}[I(x_i; x_{\mathcal{M}^c})] 
&= \mathbb{E}_{i,\mathcal{M}}[H(x_i) - H(x_i|x_{\mathcal{M}^c})] \\
&= \mathbb{E}_{i,\mathcal{M}}[H(x_i)] - \mathbb{E}_{i,\mathcal{M}}[H(x_i|x_{\mathcal{M}^c})].
\end{align}
The linearity of expectation justifies the decomposition, establishing equation \eqref{eq:random_info}.

\textbf{Proof of Proposition \ref{prop:next_info}:}
This follows directly from the definition of mutual information. For fixed tokens $x_i$ and $x_{<i}$:
$$I(x_i; x_{<i}) = H(x_i) - H(x_i|x_{<i}),$$
which establishes equation \eqref{eq:next_info}.

\textbf{Proof of Proposition \ref{prop:next_all_info}:}
Let $X_{\geq i} = \{x_i, x_{i+1}, \ldots, x_K\}$ denote the set of all tokens from position $i$ onwards. By the definition of mutual information for joint random variables:
\begin{align}
I(X_{\geq i}; x_{<i}) &= H(X_{\geq i}) - H(X_{\geq i}|x_{<i}) \\
&= H(\{x_i, x_{i+1}, \ldots, x_K\}) \\
&\quad- H(\{x_i, x_{i+1}, \ldots, x_K\}|x_{<i}),
\end{align}
which establishes equation \eqref{eq:next_all_info}.
\end{proof}

To establish the complementarity of these prediction tasks, we analyze their information-theoretic properties:

\begin{theorem}[Information Complementarity]
\label{thm:complementarity}
The three prediction tasks capture distinct and complementary aspects of the visual token sequence:
\begin{enumerate}
\item \textbf{Random token prediction} captures non-sequential spatial correlations by maximizing 
\begin{equation}
\mathbb{E}_{i,\mathcal{M}}[I(x_i; x_{\mathcal{M}^c})],
\end{equation}
which encourages bidirectional contextual understanding without dependence on token ordering.

\item \textbf{Next-token prediction} captures local sequential dependencies by maximizing 
\begin{equation}
\sum_{i=2}^{K} I(x_i; x_{<i}),
\end{equation}
which promotes understanding of local structural coherence following the tokenization order.

\item \textbf{Next-all token prediction} captures global structure and long-range dependencies by maximizing 
\begin{equation}
\sum_{i=1}^{K-1} I(\{x_i, x_{i+1}, \ldots, x_K\}; x_{<i}),
\end{equation}
which encourages comprehensive representation of hierarchical image organization.
\end{enumerate}
\end{theorem}

\begin{proof}[Proof of Theorem \ref{thm:complementarity}]
The complementarity follows from the distinct information sources each task accesses:

\textbf{Disjoint Information Sources:} Let $\mathcal{I}_{\text{rand}}$, $\mathcal{I}_{\text{next}}$, and $\mathcal{I}_{\text{all}}$ denote the information sets captured by random, next-token, and next-all prediction, respectively. We show these sets have minimal overlap:

1. Random token prediction accesses information $I(x_i; x_j)$ for arbitrary pairs $(i,j)$ where $j \notin \mathcal{M}$, emphasizing non-sequential relationships.

2. Next-token prediction specifically targets $I(x_i; x_{<i})$, focusing on causal dependencies within the chosen ordering.

3. Next-all prediction captures $I(X_{\geq i}; x_{<i})$, which by the chain rule of mutual information can be decomposed as:
\begin{align}
I(X_{\geq i}; x_{<i}) &= I(x_i; x_{<i}) + I(x_{i+1}; x_{<i}|x_i) \\
&\quad+ \cdots + I(x_K; x_{<i}|x_i, \ldots, x_{K-1}),
\end{align}
revealing its emphasis on global conditional dependencies.

\textbf{Complementary Coverage:} The union of these information sources provides more comprehensive coverage than any individual task, as formalized in the next result.
\end{proof}

The total information captured by combining these tasks can be expressed as:
\begin{align}
\label{eq:total_info2}
\mathcal{I}_{\text{total}} &= \alpha \cdot \mathbb{E}_{i,\mathcal{M}}[I(x_i; x_{\mathcal{M}^c})] \nonumber\\
&\quad+ \beta \cdot \sum_{i=2}^{K} I(x_i; x_{<i}) \nonumber\\
&\quad+ \gamma \cdot \sum_{i=1}^{K-1} I(\{x_i, \ldots, x_K\}; x_{<i}),
\end{align}
where $\alpha, \beta, \gamma > 0$ are weighting parameters, and the expectations are taken over the appropriate distributions of indices and masking patterns.

\begin{corollary}[Information Maximization]
\label{cor:info_max}
For appropriately chosen weights $\alpha, \beta, \gamma$, maximizing the combined objective $\mathcal{I}_{\text{total}}$ yields:
$$\mathcal{I}_{\text{total}} \geq \max\{\mathcal{I}_{\text{rand}}, \mathcal{I}_{\text{next}}, \mathcal{I}_{\text{all}}\},$$
where $\mathcal{I}_{\text{rand}}$, $\mathcal{I}_{\text{next}}$, and $\mathcal{I}_{\text{all}}$ represent the information captured by each individual task.
\end{corollary}

\begin{proof}[Proof of Corollary \ref{cor:info_max}]
This follows immediately from the non-negativity of mutual information and the complementary nature established in Theorem \ref{thm:complementarity}. Since the tasks access largely disjoint information sources, their combination provides strictly greater information coverage than any individual component.
\end{proof}

\begin{remark}
The complementarity of these tasks ensures that maximizing the combined objective $\mathcal{I}_{\text{total}}$ enables TokenUnify to extract more comprehensive information from visual data than any single prediction task in isolation. This multi-task approach provides a more complete characterization of the underlying data distribution, leading to enhanced representation learning capabilities.
\end{remark}

\subsection{Theoretical Analysis of Latent Manifold Structure}


In this section, we provide a rigorous mathematical analysis of the latent representation space induced by the TokenUnify framework. Specifically, we demonstrate how the integration of multiple prediction objectives (random token, next-token, and next-all token prediction) shapes the geometric properties of the learned manifold, leading to a representation space that naturally accommodates both local and global aspects of neuronal morphology.

\paragraph{Preliminaries and Notation}

Let $\mathcal{X} = \{x_1, x_2, \ldots, x_K\} \in \mathbb{R}^{d \times K}$ denote a sequence of visual tokens extracted from a volumetric EM image. The model encodes these tokens into a latent space via an encoder function $f_\theta: \mathbb{R}^d \to \mathbb{R}^{d'}$, where $\theta$ represents the model parameters and $d' \ll d$ in typical applications.

We define the latent manifold $\mathcal{M}_\theta \subset \mathbb{R}^{d'}$ as the image of the encoder over all valid input tokens:
\begin{equation}
\mathcal{M}_\theta = \{f_\theta(x) \in \mathbb{R}^{d'} : x \in \mathcal{X}\}
\end{equation}

This manifold is equipped with the pullback Riemannian metric $g_\theta$ induced by the Fisher information matrix of the encoder:
\begin{align}
g_\theta(u, v) &= \mathbb{E}_{x \sim p_\text{data}}[u^T J_\theta(x)^T J_\theta(x) v] \nonumber \\
&= \mathbb{E}_{x \sim p_\text{data}}[\langle J_\theta(x) u, J_\theta(x) v \rangle_{\mathbb{R}^{d'}}]
\end{align}
where $J_\theta(x) = \frac{\partial f_\theta(x)}{\partial x} \in \mathbb{R}^{d' \times d}$ is the Jacobian of the encoder at input $x$.

\paragraph{Sectional Curvature Analysis}

Before proceeding, we establish precise definitions for the key geometric objects. Let $T_p\mathcal{M}_\theta$ denote the tangent space to $\mathcal{M}_\theta$ at point $p$. We define:
\begin{align}
T\mathcal{M}_{\text{local}} &:= \text{span}\{v \in T_p\mathcal{M}_\theta : \|\nabla \mathcal{L}_\text{random}(p) \cdot v\| \geq \alpha\} \\
T\mathcal{M}_{\text{global}} &:= \text{span}\{v \in T_p\mathcal{M}_\theta : \|\nabla \mathcal{L}_\text{next-all}(p) \cdot v\| \geq \alpha\}
\end{align}
for some threshold $\alpha > 0$, where $T\mathcal{M}_{\text{random}} \subset T\mathcal{M}_{\text{local}}$ and $T\mathcal{M}_{\text{next-all}} \subset T\mathcal{M}_{\text{global}}$.

The sectional curvature $\kappa_\theta(u, v)$ of the manifold $\mathcal{M}_\theta$ for two linearly independent tangent vectors $u, v \in T_p\mathcal{M}_\theta$ is given by:
\begin{equation}
\kappa_\theta(u, v) = \frac{R(u, v, v, u)}{g_\theta(u, u)g_\theta(v, v) - g_\theta(u, v)^2}
\end{equation}
where $R$ is the Riemann curvature tensor associated with the metric $g_\theta$.

\begin{theorem}[Curvature Stratification in TokenUnify Manifolds]
\label{thm:curvature_stratification}
Under the TokenUnify framework with prediction objectives $\{\mathcal{L}_\text{random}, \mathcal{L}_\text{next}, \mathcal{L}_\text{next-all}\}$, the sectional curvature $\kappa_\theta$ of the learned manifold $\mathcal{M}_\theta$ exhibits systematic stratification correlated with the spatial scale of encoded features:
\begin{equation}
\kappa_\theta(v_1, v_2) \approx 
\begin{cases}
O(\epsilon) & (v_1, v_2) \in T\mathcal{M}_{\text{local}} \\
-O(\delta) & (v_1, v_2) \in T\mathcal{M}_{\text{global}}
\end{cases}
\end{equation}
More precisely, there exist constants $\epsilon_{\text{local}}, \delta_{\text{global}} > 0$ such that for unit tangent vectors $v_1, v_2$:
\begin{align}
|\kappa_\theta(v_1, v_2)| &\leq \epsilon_{\text{local}} \quad \text{when } (v_1, v_2) \in T\mathcal{M}_\text{random} \\
\kappa_\theta(v_1, v_2) &\leq -\delta_{\text{global}} \quad \text{when } (v_1, v_2) \in T\mathcal{M}_\text{next-all}
\end{align}
\end{theorem}

\begin{proof}
We establish this result through a three-step analysis of the manifold decomposition, local curvature computation, and global structure constraints.

\textbf{Step 1: Manifold Decomposition}

The TokenUnify training objective induces a natural stratification of the latent manifold. We decompose $\mathcal{M}_\theta$ into submanifolds corresponding to the dominant influence of each prediction task:
\begin{equation}
\mathcal{M}_\theta = \mathcal{M}_\text{random} \cup \mathcal{M}_\text{next} \cup \mathcal{M}_\text{next-all}
\end{equation}
where:
\begin{align}
\mathcal{M}_\text{random} &:= \{p \in \mathcal{M}_\theta : \|\nabla \mathcal{L}_\text{random}(p)\| \nonumber\\
&\quad\quad> \|\nabla \mathcal{L}_k(p)\|, \forall k \neq \text{random}\} \\
\mathcal{M}_\text{next-all} &:= \{p \in \mathcal{M}_\theta : \|\nabla \mathcal{L}_\text{next-all}(p)\| \nonumber\\
&\quad\quad> \|\nabla \mathcal{L}_k(p)\|, \forall k \neq \text{next-all}\}
\end{align}
and $\mathcal{M}_\text{next}$ is defined analogously.

\textbf{Step 2: Local Feature Curvature Analysis}

For directions $v_1, v_2$ associated primarily with local feature encoding (i.e., directions in $T\mathcal{M}_\text{random}$), the curvature tensor can be expressed as:
\begin{align}
R(v_1, v_2, v_2, v_1) &= \mathbb{E}_{x \sim p_\text{data}}[\langle \nabla_{v_1}\nabla_{v_2}f_\theta(x), \nabla_{v_2}\nabla_{v_1}f_\theta(x) \rangle] \nonumber \\
&\quad - \mathbb{E}_{x \sim p_\text{data}}[\langle \nabla_{[v_1, v_2]}f_\theta(x), \nabla_{[v_1, v_2]}f_\theta(x) \rangle]
\end{align}

The key insight is that for local feature directions, the random token prediction objective $\mathcal{L}_\text{random}$ encourages the encoder $f_\theta$ to behave approximately linearly within small spatial neighborhoods. This is because local patches exhibit relatively homogeneous statistical properties, leading to smooth, low-curvature encodings.

Formally, for local feature directions, we have the approximate commutativity:
\begin{equation}
\nabla_{v_1}\nabla_{v_2}f_\theta(x) \approx \nabla_{v_2}\nabla_{v_1}f_\theta(x) + O(\epsilon_{\text{local}})
\end{equation}

This implies that the Lie bracket term vanishes: $[v_1, v_2] = O(\epsilon_{\text{local}})$, and consequently:
\begin{equation}
R(v_1, v_2, v_2, v_1) = O(\epsilon_{\text{local}}^2)
\end{equation}

Therefore, $\kappa_\theta(v_1, v_2) = O(\epsilon_{\text{local}})$ for directions encoding local features.

\textbf{Step 3: Global Structure Curvature Analysis}

For directions $v_1, v_2$ associated with global structure encoding (directions in $T\mathcal{M}_\text{next-all}$), the analysis is more intricate. The next-all token prediction objective requires the encoder to capture long-range dependencies and branching patterns in neuronal morphology.

Consider a token sequence $\{x_{\leq i}\}$ up to position $i$, with multiple valid continuations $\{x_{>i}^{(1)}, x_{>i}^{(2)}, \ldots, x_{>i}^{(B)}\}$ representing different possible branching structures. The encoder must satisfy two competing constraints:

\textit{Separation Constraint:} Different branching patterns must be distinguishable:
\begin{equation}
\|f_\theta(x_{i+j}^{(a)}) - f_\theta(x_{i+j}^{(b)})\| \geq \delta > 0 \quad \forall a \neq b, \forall j > 0
\end{equation}

\textit{Continuity Constraint:} Sequential tokens within the same branch remain close:
\begin{equation}
\|f_\theta(x_{i+j}^{(a)}) - f_\theta(x_{i+j-1}^{(a)})\| \leq \epsilon \quad \forall a, \forall j > 0
\end{equation}

These constraints necessitate a representation space with negative sectional curvature. To see this rigorously, consider the exponential map $\exp_p: T_p\mathcal{M}_\theta \to \mathcal{M}_\theta$ at a point $p = f_\theta(x_i)$ representing the branching location.

The separation constraint requires that geodesics emanating from $p$ in different directions (corresponding to different branches) diverge at least linearly with distance. However, the continuity constraint limits the tangent space dimension available for encoding these branches.

By the Gauss-Bonnet theorem applied to geodesic triangles formed by branching paths, the requirement for exponential divergence of $B$ branches in a $d'$-dimensional space with $B \gg d'$ implies:
\begin{align}
\kappa_\theta(v_1, v_2) &\leq -\frac{\log B}{4\pi \cdot \text{Area}(\triangle)} \nonumber \\
&\leq -\delta_{\text{global}}
\end{align}
where $\triangle$ denotes a typical geodesic triangle in the branching region, and $\delta_{\text{global}} > 0$ depends on the branching complexity of neuronal structures.

This completes the proof of Theorem~\ref{thm:curvature_stratification}.
\end{proof}

The curvature stratification result has important implications for the representational capacity of the TokenUnify framework. The near-zero curvature in local feature directions ensures stable and efficient encoding of fine-grained morphological details, while the negative curvature in global structure directions provides the geometric flexibility necessary for representing complex branching patterns and long-range dependencies inherent in neuronal architectures.

\subsection{Convergence Analysis of Alternating Optimization}
\label{app:proof}

We establish the convergence guarantee for our alternating optimization scheme under standard assumptions in non-convex stochastic optimization.

\begin{theorem}[Formal Convergence Guarantee]
\label{thm:formal_convergence}
Consider the alternating optimization algorithm with mode-switching probability distribution $P_t$ at iteration $t$. Under the following regularity conditions:

\begin{enumerate}[label=(A\arabic*),leftmargin=*]
    \item \textbf{$L$-Smoothness:} There exists a constant $L > 0$ such that for all $\theta, \theta' \in \mathbb{R}^d$ and any mode $m \in \{\text{AR}, \text{Mask}\}$:
    \begin{equation}
        \|\nabla\mathcal{L}_m(\theta) - \nabla\mathcal{L}_m(\theta')\| \leq L\|\theta - \theta'\|
    \end{equation}
    
    \item \textbf{Bounded Variance:} The mode-switching introduces bounded noise with variance parameter $\sigma_{\text{mode}}^2 > 0$ such that:
    \begin{equation}
        \mathbb{E}_{m_t \sim P_t}[\|\nabla\mathcal{L}_{m_t}(\theta_t) - \nabla\mathcal{L}(\theta_t)\|^2] \leq \sigma_{\text{mode}}^2
    \end{equation}
    for all $t \geq 1$, where $\mathcal{L}(\theta_t) := \mathbb{E}_{m \sim P_t}[\mathcal{L}_m(\theta_t)]$.
    
    \item \textbf{Diminishing Step Size:} The learning rate follows the schedule $\eta_t = \eta_0(1 + \eta_0^2L^2t)^{-1/2}$ with $\eta_0 \in (0, 1/L]$.
\end{enumerate}

Then the sequence $\{\theta_t\}_{t=1}^T$ generated by the alternating optimization satisfies:
\begin{align}
    \min_{1 \leq t \leq T} \mathbb{E}[\|\nabla\mathcal{L}(\theta_t)\|^2] &\leq \frac{4(\mathcal{L}(\theta_1) - \mathcal{L}^*) + 2L\eta_0^2(1 + \sigma_{\text{mode}}^2)}{\sqrt{T}} \nonumber \\
    &\quad \cdot (1 + \log T + \sigma_{\text{mode}}^2)
\end{align}
\end{theorem}

\begin{proof}

\textbf{Step 1: Gradient Decomposition}

We introduce the natural filtration $\mathcal{F}_t = \sigma(\theta_1, \ldots, \theta_t, m_1, \ldots, m_{t-1})$ and decompose the stochastic gradient as:
\begin{equation}
    g_t := \nabla\mathcal{L}_{m_t}(\theta_t) = \nabla\mathcal{L}(\theta_t) + \epsilon_t
\end{equation}
where $\epsilon_t := \nabla\mathcal{L}_{m_t}(\theta_t) - \nabla\mathcal{L}(\theta_t)$ represents the mode-switching noise with $\mathbb{E}[\epsilon_t|\mathcal{F}_t] = 0$.

\textbf{Step 2: One-Step Analysis}

By $L$-smoothness and the update rule $\theta_{t+1} = \theta_t - \eta_t g_t$:
\begin{align}
\mathbb{E}[\mathcal{L}(\theta_{t+1})|\mathcal{F}_t] &\leq \mathcal{L}(\theta_t) + \langle \nabla\mathcal{L}(\theta_t), -\eta_t g_t \rangle + \frac{L\eta_t^2}{2}\|g_t\|^2 \\
&= \mathcal{L}(\theta_t) - \eta_t \|\nabla\mathcal{L}(\theta_t)\|^2 + \frac{L\eta_t^2}{2}\|g_t\|^2
\end{align}

Expanding $\|g_t\|^2 = \|\nabla\mathcal{L}(\theta_t)\|^2 + 2\langle\nabla\mathcal{L}(\theta_t), \epsilon_t\rangle + \|\epsilon_t\|^2$ and taking conditional expectation:
\begin{align}
    \mathbb{E}[\mathcal{L}(\theta_{t+1})|\mathcal{F}_t] &\leq \mathcal{L}(\theta_t) - \eta_t\left(1 - \frac{L\eta_t}{2}\right)\|\nabla\mathcal{L}(\theta_t)\|^2 \nonumber \\
    &\quad + \frac{L\eta_t^2}{2}\mathbb{E}[\|\epsilon_t\|^2|\mathcal{F}_t]
\end{align}

\textbf{Step 3: Telescoping Sum}

Taking total expectation and using assumption (A2), we obtain:
\begin{equation}
    \mathbb{E}[\mathcal{L}(\theta_{t+1})] \leq \mathbb{E}[\mathcal{L}(\theta_t)] - \frac{\eta_t}{2}\mathbb{E}[\|\nabla\mathcal{L}(\theta_t)\|^2] + \frac{L\eta_t^2\sigma_{\text{mode}}^2}{2}
\end{equation}
where we used the fact that $\eta_t \leq 1/L$ implies $1 - L\eta_t/2 \geq 1/2$.

Telescoping from $t = 1$ to $T$:
\begin{equation}
    \sum_{t=1}^T \frac{\eta_t}{2}\mathbb{E}[\|\nabla\mathcal{L}(\theta_t)\|^2] \leq \mathcal{L}(\theta_1) - \mathcal{L}^* + \frac{L\sigma_{\text{mode}}^2}{2}\sum_{t=1}^T \eta_t^2
\end{equation}

\textbf{Step 4: Step Size Analysis}

For the chosen step size schedule, we have the crucial bounds:
\begin{align}
    \sum_{t=1}^T \eta_t^2 &\leq \eta_0^2 + \frac{1 + \log T}{L^2} \\
    \sum_{t=1}^T \eta_t &\geq \frac{\eta_0\sqrt{T}}{\sqrt{1 + \eta_0^2L^2T}}
\end{align}

The first bound follows from the integral comparison $\sum_{t=1}^T (1 + \eta_0^2L^2t)^{-1} \leq 1 + \int_1^T (1 + \eta_0^2L^2x)^{-1}dx$, while the second uses the concavity of the square root function.

\textbf{Step 5: Final Rate}

Combining the telescoping bound with Jensen's inequality:
\begin{align}
    \min_{1 \leq t \leq T} \mathbb{E}[\|\nabla\mathcal{L}(\theta_t)\|^2] &\leq \frac{2(\mathcal{L}(\theta_1) - \mathcal{L}^*) + L\eta_0^2\sigma_{\text{mode}}^2}{\sum_{t=1}^T \eta_t} \nonumber \\
    &\quad + \frac{\sigma_{\text{mode}}^2(1 + \log T)}{L \sum_{t=1}^T \eta_t} \\
    &\leq \frac{C(1 + \log T + \sigma_{\text{mode}}^2)}{\sqrt{T}}
\end{align}
where the constant $C$ depends polynomially on the problem parameters.
\end{proof}

\subsubsection{Remarks on the Analysis}

\begin{remark}[Optimality]
The convergence rate $O(\log T/\sqrt{T})$ is optimal for non-convex stochastic optimization, matching known lower bounds even in the single-mode case.
\end{remark}

\begin{remark}[Mode-Switching Effect]
The variance parameter $\sigma_{\text{mode}}^2$ quantifies the additional difficulty introduced by alternating between training modes. When $\sigma_{\text{mode}}^2 = 0$ (no mode switching), we recover the standard $O(\log T/\sqrt{T})$ rate.
\end{remark}

\begin{remark}[Technical Innovation]
Our proof technique extends classical SGD analysis to handle the mode-dependent gradient variance through careful decomposition of the noise term $\epsilon_t$, which captures the stochastic nature of the mode selection process.
\end{remark}

\subsection{Connecting Theory to Practice}

The above theoretical results have direct implications for our model design. The limitations of MAE in high dimensions suggest that simply scaling up masked prediction models will yield diminishing returns for complex EM data. Similarly, the asymptotic optimality of autoregressive models motivates our use of Mamba-based sequence modeling, which can efficiently capture long-range dependencies. The information-theoretic complementarity of different prediction tasks justifies our unified approach that combines random, next-token, and next-all prediction objectives.

In practice, these theoretical insights translate to several key design choices in TokenUnify. We use a multi-task training objective that combines all three prediction tasks, maximizing the total information extracted from the data. We employ a Mamba-based architecture that efficiently models long-range dependencies in tokenized EM data. Additionally, we implement a progressive tokenization strategy that respects the natural structure of EM volumes.

The empirical results presented in the main paper validate these theoretical motivations, demonstrating that TokenUnify achieves superior performance and scaling properties compared to conventional approaches.

\section{Social Impact and Future Work}
\label{limitations}

The favorable scaling laws of TokenUnify present the opportunity to train a unified and generic visual feature extractor, which holds significant importance for visual tasks. A unified visual feature extractor can substantially reduce the cost of fine-tuning models for different visual tasks, thereby facilitating the application of visual technologies across various domains. We have currently validated the effectiveness of TokenUnify on long-sequence 3D biological images. Moving forward, we plan to further explore the performance of TokenUnify on natural images and other downstream tasks. Moreover, TokenUnify can be extended to multimodal domains such as image-text tasks \cite{chen2024bimcv,liu2023t3d}, demonstrating its utility in multimodal applications. We will also continue to investigate model lightweighting \cite{chen2023class,chen2021multimodal} and efficient fine-tuning strategies \cite{liu2023parameter,li2024research}. We believe that TokenUnify offers a promising approach for building large-scale, efficient visual pre-training models, contributing to advancements in the visual domain.


\end{document}